\documentclass{article}
\usepackage{fix-cm}
\usepackage{iclr2027_conference,times}

\usepackage{amsmath,amsfonts,bm}

\def\eqref#1{equation~\ref{#1}}

\def\1{\bm{1}}

\def\vx{{\bm{x}}}
\def\vy{{\bm{y}}}

\DeclareMathAlphabet{\mathsfit}{\encodingdefault}{\sfdefault}{m}{sl}
\SetMathAlphabet{\mathsfit}{bold}{\encodingdefault}{\sfdefault}{bx}{n}

\usepackage{xurl}
\usepackage{url}
\usepackage{graphicx}
\usepackage{amssymb}
\usepackage{booktabs}
\usepackage{array}
\usepackage{listings}
\usepackage{tikz}
\usepackage{placeins}
\usepackage{float}
\usepackage{xcolor}
\usepackage{colortbl}

\usetikzlibrary{arrows.meta,positioning,bending,fit} 

\definecolor{ink}{HTML}{222831}           
\definecolor{gridink}{HTML}{D9DCE1}       
\definecolor{demandink}{HTML}{2D3F66}     
\definecolor{supplyslate}{HTML}{007866}   
\definecolor{prioritywine}{HTML}{AC3828}  
\definecolor{knowledgeink}{HTML}{007866}  
\definecolor{warmneg}{HTML}{5E4835}       
\definecolor{criterionhead}{HTML}{2D3F66} 
\definecolor{criterionfavor}{HTML}{007866}
\definecolor{criterionavoid}{HTML}{AC3828}

\definecolor{methodink}{HTML}{222222}
\definecolor{methodblue}{HTML}{D8E5FA}   
\definecolor{methodpink}{HTML}{F4D9D6}   
\definecolor{methodcream}{HTML}{FFF3CE} 
\definecolor{methodgreen}{HTML}{DEECD7} 
\definecolor{methodborder}{HTML}{86AC70}
\definecolor{methodpurple}{HTML}{E5DCEA}
\definecolor{methodgroup}{HTML}{F8F8F8} 
\definecolor{methodgray}{HTML}{959595}  

\colorlet{tablehead}{demandink!10}
\colorlet{tablegroup}{ink!5}
\colorlet{tablerubric}{supplyslate!9}
\colorlet{tableflag}{prioritywine!11}
\newcommand{\tablestyle}{\renewcommand{\arraystretch}{1.16}}
\newcommand{\tablepanel}[2]{%
  \rowcolor{tablegroup}\multicolumn{#1}{l}{\strut\textbf{#2}}\\}

\newcommand{\Pbase}{\mbox{\textsc{Instruct}}}
\newcommand{\Pdpo}{\mbox{DPO}}
\newcommand{\Pone}{\mbox{\textsc{RubricOpt}-R1}}
\newcommand{\Ptwo}{\mbox{\textsc{RubricOpt}-R2}}
\newcommand{\PtwoK}{\mbox{\textsc{RubricOpt}-R2\,(\textsc{keep})}}
\newcommand{\PtwoR}{\mbox{\textsc{RubricOpt}-R2\,(\textsc{refine})}}

\usepackage[
    unicode, naturalnames, colorlinks=true, linkcolor=red!80!white, citecolor=green!60!black, urlcolor=purple!90!black ]{hyperref}

\usepackage{cleveref}

\crefname{figure}{Fig.}{Figs.} \Crefname{figure}{Fig.}{Figs.} \crefname{table}{Tab.}{Tabs.} \Crefname{table}{Tab.}{Tabs.} \crefname{section}{Sec.}{Secs.} \Crefname{section}{Sec.}{Secs.} \crefname{equation}{Eq.}{Eqs.} \Crefname{equation}{Eq.}{Eqs.} \crefname{proposition}{Prop.}{Props.} \Crefname{proposition}{Prop.}{Props.}

\newcommand{\appref}[1]{appendix~\labelcref{#1}}
\newcommand{\Appref}[1]{Appendix~\labelcref{#1}}

\newif\ifappendixcontents
\let\SavedAddcontentsline\addcontentsline
\renewcommand{\addcontentsline}[3]{%
  \ifappendixcontents\SavedAddcontentsline{#1}{#2}{#3}\fi}

\newcommand{\emptycrit}{\varnothing}
\newtheorem{proposition}{Proposition}

\title{RubricOpt: Self-Revision of Semantic Rubrics for Preference Optimization}
\iclrfinalcopy
\author{%
\bfseries Wei Chen\textsuperscript{1}\thanks{\texttt{maweichen@mail.scut.edu.cn}}\hspace{.6em}\quad Yubing Wu\textsuperscript{1}\quad Shigui Li\textsuperscript{1}\quad Delu Zeng\textsuperscript{1}\thanks{Corresponding author. \texttt{dlzeng@scut.edu.cn}}\hspace{.6em}\quad Jingfeng Liu\textsuperscript{2}\\
\bfseries Junmei Yang\textsuperscript{1}\quad Zhiheng Zhou\textsuperscript{1}\quad Bo Li\textsuperscript{1}\quad Min Chen\textsuperscript{1}\quad Zhou Wang\textsuperscript{3}\\
\textsuperscript{1}South China University of Technology \ \textsuperscript{2}LinkSprite \ \textsuperscript{3}University of Waterloo}

\begin{document}
\maketitle

\begin{abstract}
A model's failures can reveal gaps in the behavioral criteria guiding its training.
We introduce RubricOpt, a framework that lets a policy revise these criteria within preference optimization.
The policy diagnoses its own rubric-free outputs, retains requirements already covered by the rubric, and edits clauses that leave recurring failures unaddressed.
A shared semantic representation connects task demands, preference-data contrasts, and rubric clauses, while a deterministic gate checks evidence references and fixed semantic anchors.
The rubric then guides Direct Preference Optimization (DPO) during training, with preference pairs, labels, and the objective held fixed, and is removed at evaluation.
Our analysis relates the local effect on rubric-free success to semantic alignment and a residual beyond the measured factors.
Experiments on capability and open-ended generation benchmarks show that rubric conditioning can improve outcomes over ordinary DPO, with gains and regressions varying across models and datasets.
A matched continuation isolates the contribution of revision from additional training and finds mixed incremental effects for the tested edit, whose feedback overlaps the evaluation set.
RubricOpt thus brings the behavioral specification itself into the policy's feedback loop.
\end{abstract}

\section{Introduction}
Language models can supply rewards for their own iterative preference training~\citep{yuan2024selfrewarding}.
They can also improve their judgment skills by assessing their own judgments~\citep{wu2025metarewarding}.
The corresponding question for semantic training rubrics is how revising their behavioral requirements affects learned behavior, including when a revision yields gains beyond continuing training with the existing rubric.

Written rubrics specify desired response properties and can supply reward signals~\citep{gunjal2025rar} or guide the construction of synthetic preference data~\citep{gallego2025cpt}.
EvoRubrics jointly trains a policy and a rubric generator to adapt rubric-based rewards as learning proceeds~\citep{ding2026evorubrics}.
SERPO co-evolves response evidence, query-specific rubrics, and policy parameters, using rubric scores as rewards for test-time reinforcement learning~\citep{wang2026serpo}.
We study rubric revision in Direct Preference Optimization (DPO)~\citep{rafailov2023dpo}, holding the pairs, labels, and objective fixed while changing the rubric context.
A matched continuation with the existing rubric measures the edit's incremental contribution.

\begin{figure}[t]
\centering
\includegraphics[width=\linewidth]{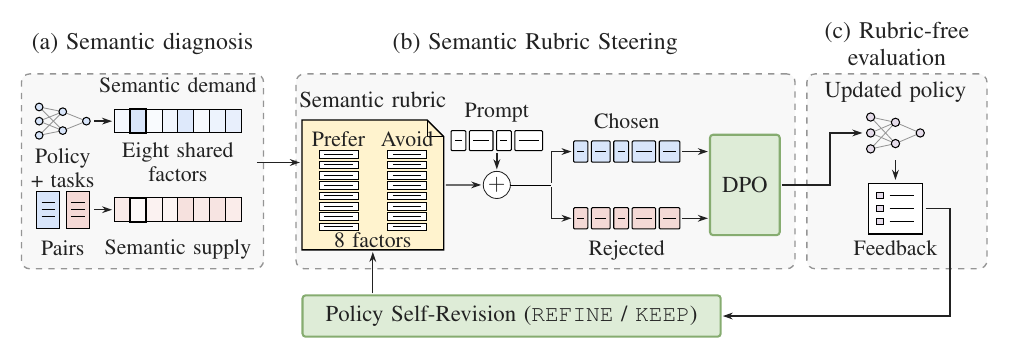}
\vspace{-8mm}
\caption{The RubricOpt loop. (a)~Task demand and preference-data supply share eight factor axes. (b)~Matched Prefer/Avoid clauses condition both responses during DPO. (c)~Rubric-free evaluation supplies feedback for policy self-revision and the next \textsc{refine}/\textsc{keep} decision. Dashed frames group stages. Green boxes denote computation.}
\label{fig:overview}
\end{figure}

We introduce RubricOpt to study how a model revises its own training rubric across preference-training rounds (\cref{fig:overview}).
Fine-grained evaluation describes response quality through explicit skills~\citep{ye2023flask}.
We use a fixed eight-factor ontology to connect the model's task failures, the contrasts in its preference data, and the rubric's Prefer/Avoid clauses.
Diagnosis-guided self-evolution uses failure histories to direct subsequent learning~\citep{wei2026diagevo}.
RubricOpt applies this diagnostic idea to rubric revision, with the current policy reading task outcomes, self-judged factor scores, and judge feedback on its own outputs to attribute recurring failures to the eight factors.
It retains explicit requirements for further learning and proposes clause edits where the wording leaves recurring failures uncovered.
A deterministic gate checks evidence references, factor anchors, and edit size before carrying the rubric into the next round.
Training-only Semantic Rubric Steering uses the resulting rubric to condition both responses of a preference pair during DPO~\citep{rafailov2023dpo}, with the pairs, labels, and objective held fixed.
Its exposure decreases during training, and evaluation uses the resulting policy without the rubric.
This use of training-only information has a precedent in privileged-information learning~\citep{vapnik2009lupi}.
Our analysis relates the extra update induced by a rubric to rubric-free success through its alignment with task demand (\cref{sec:transfer}).

The experiments evaluate how changes to the training specification affect comprehensive capability and open-ended generation.
Fixed-state probes and criterion comparisons show that rubric text changes the learning update and can alter rubric-free outcomes (\cref{sec:context_readouts}).
Across model families and preference corpora, rubric conditioning improves some capability and generation evaluations over ordinary DPO, with gains and regressions varying by setting (\cref{tab:main_results,tab:mistral_results}).
The dataset comparison relates these effect profiles to the behavioral contrasts available in each corpus (\cref{sec:dataset_effects}).
A matched continuation separates revision from further optimization and finds mixed incremental results from the tested edit (\cref{tab:continuation_main}).

RubricOpt makes the behavioral specification itself an object of policy-driven revision within preference learning.
Shared semantic coordinates connect failure diagnosis and preference-data contrasts to editable rubric clauses, while training-only steering connects those clauses to subsequent parameter updates.
This closes the loop from learned behavior to the specification of further learning, with the effect of revision evaluated independently of additional training.

\section{Related Work}

\paragraph{Preference Optimization and Credit Assignment.}
The family descends from RLHF, in which a reward model is fit to pairwise human judgements~\citep{stiennon2020learning} and a policy is then optimised against it at assistant scale~\citep{ouyang2022training,bai2022harmless}.
DPO instead fits chosen--rejected pairs with a classification objective~\citep{rafailov2023dpo}, whose implicit reward is the log density ratio between the policy and the reference~\citep{sugiyama2012density,chen2025dequantified}.
Later objectives change one component of that recipe: IPO replaces it with a squared loss on the pairwise margin~\citep{azar2024ipo}, KTO drops the pairing requirement by scoring each response against a reference point~\citep{ethayarajh2024kto}, and SimPO removes the reference model and normalises the margin by length~\citep{meng2024simpo}.
Two analyses place the family in a wider frame: one connects these objectives to imitation learning~\citep{xiao2025dil}, and one disentangles how their update suppresses the rejected response and preserves the chosen one~\citep{chen2026towards}.
A complementary line assigns credit inside a response rather than to the response as a whole: TDPO reweights tokens by a forward-KL penalty~\citep{zeng2024tdpo}, TIS-DPO reweights them by importance sampling against an estimated weight~\citep{liu2024tisdpo}, SEPO selects tokens using a token-level reward estimate~\citep{yang2024sepo}, T-REG uses a token-level reward as a regulariser~\citep{zhou2025treg}, and TIDPO guides the update by token importance~\citep{yang2025tidpo}.
RubricOpt operates at a different level: a semantic factor names a behavioral distinction that can span several tokens or reasoning steps, while the pairwise loss itself is left unchanged.

\paragraph{Criteria and Behavioral Diagnosis.}
Written criteria already enter learning through several interfaces.
As reward signals, Rubrics as Rewards scores on-policy responses against a rubric~\citep{gunjal2025rar}.
As data, Configurable Preference Tuning builds rubric-conditioned synthetic preference pairs~\citep{gallego2025cpt}.
And as weights, CoRT contrasts rubric-conditioned with rubric-free replays to form token-level weights~\citep{zhang2026cort}.
Criterion-conditioned training also has close relatives outside preference optimisation: learning using privileged information trains on a signal that is absent at test time~\citep{vapnik2009lupi}.
Context distillation trains on a context and then evaluates without it, introduced as an alignment technique~\citep{askell2021general} and developed as a standalone method~\citep{snell2022context}.
And classifier-free guidance interpolates between conditioned and unconditioned predictions~\citep{ho2022cfg}.
On the diagnostic side, fine-grained evaluation describes response quality through interpretable skill sets and error taxonomies---FLASK by grading along skill dimensions~\citep{ye2023flask}, EvalTree by profiling weaknesses as a hierarchical capability tree~\citep{zeng2025evaltree}, and ErrorMap and ErrorAtlas by charting the failure landscape into error categories ~\citep{ashury2026erroratlas}---and diagnosis-driven post-training then acts on what those tools find, with DiagEvo feeding diagnosed errors back into self-evolution through a hierarchical error memory~\citep{wei2026diagevo} and APTER turning expert-grounded rubrics into post-training supervision ~\citep{wang2026apter}.
RubricOpt uses shared semantic coordinates to connect task diagnosis, preference-data contrasts, and edits to the training rubric.

\paragraph{Self-Improvement and Recursive Feedback.}
Self-improvement methods differ in the component they revise and how that revision enters subsequent behavior.
Self-Refine edits outputs from self-feedback~\citep{madaan2023selfrefine}, and automatic prompt optimisation edits inference-time text from feedback~\citep{pryzant2023automatic}.
In parameter learning, Self-Rewarding uses the model to generate its own reward signals~\citep{yuan2024selfrewarding}, Meta-Rewarding trains judgment skills through a meta-judge~\citep{wu2025metarewarding}, and self-play preference optimisation regenerates preferences~\citep{wu2025sppo}.
Rubric evolution also enters the reward interface through joint policy--generator learning~\citep{ding2026evorubrics} and co-evolving query-specific rubrics during test-time reinforcement learning~\citep{wang2026serpo}.
RubricOpt places self-revision in the semantic rubric that conditions preference training.
The current policy proposes clause edits from feedback about its own behavior, then learns under the resulting rubric in the next round.
The initial clause bank, factor ontology, task outcomes, and fixed scoring prompt provide the reference points for revision.

\section{RubricOpt: Method and Analysis}
RubricOpt alternates preference learning under a written rubric with revision of that rubric by the resulting policy.

\paragraph{Problem Setup.}
\label{sec:setup_notation}
We study preference optimization on a fixed dataset $\mathcal D=\{(\vx,\vy^{+},\vy^{-})\}$ of chosen--rejected pairs, with a trainable policy $\pi_\theta$ and a frozen reference $\pi_{\mathrm{ref}}$ trained by DPO~\citep{rafailov2023dpo},
\begin{equation}
\mathcal L=-\mathbb E\left[\log\sigma\left(\beta\left[
\bar\ell_\theta(\vy^{+}\mid\vx)-\bar\ell_{\mathrm{ref}}(\vy^{+}\mid\vx)
-\bar\ell_\theta(\vy^{-}\mid\vx)+\bar\ell_{\mathrm{ref}}(\vy^{-}\mid\vx)
\right]\right)\right],
\end{equation}
with $\bar\ell(\vy\mid\vx)=L^{-1}\log\pi(\vy\mid\vx)$ averaged over the $L$ response tokens.
The pairs, labels, and objective are held fixed; only the text that conditions a pair may change, which is the degree of freedom RubricOpt introduces.

Behavior is described by a fixed eight-factor ontology $\mathcal F=\{f_1,\ldots,f_8\}$ (\appref{sec:ontology}), on which a blind judge returns one ordinal score per factor, $z_k(\vx,\vy)\in\{0,\ldots,4\}$ (\appref{sec:judge}).
Every behavioral quantity here is a contrast between two groups of these scores, and only two contrasts are used.
Responses to a benchmark $B$ are grouped by the task outcome $o\in\{0,1\}$, and \emph{semantic demand} is the mean score gap between the groups, measured for one model on one task,
\begin{equation}
D_{m,B,k}=\mathbb E[z_k\mid o=1,m,B]-\mathbb E[z_k\mid o=0,m,B].
\label{eq:demand}
\end{equation}
Inside a preference pair the two responses are contrasted by $\Delta z_k=z_k(\vx,\vy^{+})-z_k(\vx,\vy^{-})$, and \emph{semantic supply} summarizes its distribution,
\begin{equation}
\mu_k=\mathbb E[\Delta z_k],\qquad
C_k=\Pr(\Delta z_k\ne0),\qquad
P_k=\Pr(\Delta z_k>0),
\label{eq:supply}
\end{equation}
as the mean signed contrast, the contrast coverage, and the share of pairs whose contrast agrees with the label.
Demand is a property of the model--task pair and supply a property of the dataset, and both live on the same eight axes.

The artifact RubricOpt edits is a \emph{rubric} $R_r$, a short natural-language specification whose clauses are keyed to factors, with $r$ indexing rounds and a frozen clause bank supplying the initial $R_0$.
For factor group $\mathcal G\subseteq\mathcal F$ the training context is
\begin{equation}
c_r=\operatorname{render}(R_r,\mathcal G),
\label{eq:rubric_context}
\end{equation}
where the renderer concatenates the current Prefer/Avoid clauses in the fixed factor order and $\emptycrit$ denotes the absent criterion.
Here $\mathcal G=\mathcal F$ except in the factor-set ablation of \cref{sec:ablation}.
A conditioning context is therefore $u\in\{c_r,\emptycrit\}$, $\pi_\theta(\vy\mid\vx,u)$ is the policy under that context, the main evaluations use $\pi_\theta(\vy\mid\vx)$, and \cref{sec:weights} tests rubric reinsertion at inference time.
\Appref{sec:notation} collects the symbols used throughout.

\subsection{Diagnosis and Conditioning}
\paragraph{Shared Behavioral Coordinates.}
\label{sec:diagnosis}
With the coordinates of \cref{sec:setup_notation} fixed, the diagnosis is a measurement on one model and one dataset: we score responses from Qwen3-4B on each benchmark and each HelpSteer3 pair, then read out demand $D_{m,B,k}$ (\cref{eq:demand}) and supply $\mu_k$, $C_k$, $P_k$ (\cref{eq:supply}).
We mark demand gaps of at least $0.15$ on the $0$--$4$ scale as meaningful and contrast coverage below $0.25$ as sparse.
\Cref{tab:demand_detail} reports the complete Qwen3-4B demand matrix. Judge checks and length controls appear in appendices~\labelcref{sec:judge,sec:demand_detail}.

The measurement separates the factors into Common ($f_2$ reasoning validity, $f_5$ completeness), Specific ($f_4$ factual grounding, $f_7$ clarity, $f_8$ efficiency), and Low-demand ($f_1$ comprehension, $f_3$ verification, $f_6$ instruction adherence).
The shared factor ontology links task diagnosis, preference-data contrasts, and clause-level feedback.
All eight factors enter the initial rubric from the frozen clause bank.
The demand and supply measurements characterize the learning setting; recurring failures whose requirements are incompletely expressed determine subsequent wording edits.

\paragraph{Training-Only Conditioning.}
Each round fixes $R_r$ and its full-factor context $c_r=\operatorname{render}(R_r,\mathcal F)$.
At step $t$ of $N$, let $\rho_t=t/N$ and independently sample $u_t=c_r$ with probability $q(\rho_t)$ and $u_t=\emptycrit$ otherwise for each pair, sharing it across both responses, the policy, and the frozen reference.
Exposure is $q(\rho)=1$ for $\rho\le 0.1$, $\{1+\cos[\pi(\rho-0.1)/0.8]\}/2$ for $0.1<\rho\le 0.9$, and $0$ thereafter.
With response-token mean log-probability $\bar\ell(\vy\mid\vx,u)=L(\vy)^{-1}\log\pi(\vy\mid\vx,u)$, training uses the unchanged DPO objective
\begin{equation}
\begin{aligned}
M_u&=\bar\ell_\theta(\vy^+\mid\vx,u)-\bar\ell_{\rm ref}(\vy^+\mid\vx,u)
-\bar\ell_\theta(\vy^-\mid\vx,u)+\bar\ell_{\rm ref}(\vy^-\mid\vx,u),
\\
\mathcal L&=-\mathbb E[\log\sigma(\beta M_{u_t})].
\end{aligned}
\label{eq:objective}
\end{equation}
Rubric tokens are masked from the loss and excluded from response-length normalization.
Evaluation uses $\pi_\theta(\vy\mid\vx)$ without a rubric.

\subsection{Evidence-Grounded Recursive Refinement}
Given rubric $R_r$, the policy trained with its rendered context $c_r$, and item-level benchmark feedback $\mathcal B_{r+1}$, one iteration is
\begin{equation}
R_r\longrightarrow c_r\longrightarrow\pi_{r+1}\longrightarrow \mathcal B_{r+1}
\longrightarrow R_{r+1}.
\label{eq:rubricopt}
\end{equation}
Five prompt stages extract benchmark failures, synthesize recurring patterns, attribute them to the eight factors, propose conservative clause updates, and audit the candidate rubric.
The policy $\pi_{r+1}$ performs this analysis and proposes the edits itself, using task outcomes, self-judged factor scores, and fixed-judge feedback on its own rubric-free outputs.
Attribution compares each recurring failure with the current clauses. An explicit requirement is retained for further learning, while an incompletely expressed requirement is eligible for refinement.
A separate deterministic gate checks evidence references, factor anchors, and edit size before accepting a change.
A covered requirement or a failed check returns \textsc{keep}.
The accepted rubric is rendered as $c_{r+1}=\operatorname{render}(R_{r+1},\mathcal G)$ for the next training round; \textsc{keep} retains the corresponding clauses from $R_r$.
Full prompts, stage contracts, and the deterministic acceptance checks are in \appref{sec:workflow_full}.

\begin{figure}[!ht]
\centering
\includegraphics[width=\linewidth]{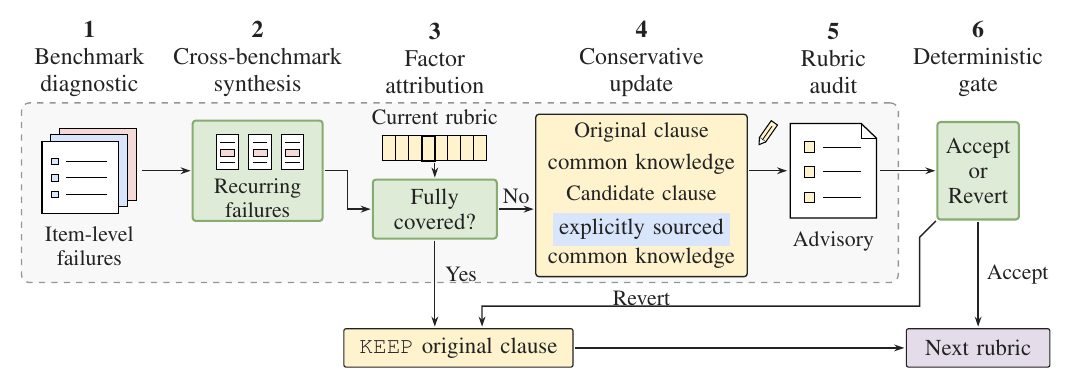}
\vspace{-8mm}
\caption{Policy-driven rubric refinement (\appref{sec:workflow_full}). Prompt stages 1--5 diagnose failures, attribute gaps, propose edits, and audit the candidate. The audit is advisory; step 6 decides acceptance. Covered requirements and rejected edits retain the original clause. The excerpt shows the factual-grounding edit. The dashed frame groups prompt stages. Green boxes denote computation.}
\label{fig:rubricopt}
\end{figure}

\subsection{Analysis of the Steering Effect}
\paragraph{The Update and Its Rubric-Free Effect.}
\label{sec:transfer}
Fix one round, with rubric context $c=c_r$ and absent context $0=\emptycrit$.
For $s_u(\vy)=L^{-1}\nabla_\theta\log\pi_\theta(\vy\mid\vx,u)$, the per-pair loss gradient is
\begin{equation}
 g_u=\nabla_\theta[-\log\sigma(\beta M_u)]
 =-\kappa_u[s_u(\vy^+)-s_u(\vy^-)],\qquad
 \kappa_u=\beta\sigma(-\beta M_u),
 \label{eq:decomposition}
\end{equation}
Let $d_u=s_u(\vy^+)-s_u(\vy^-)$ and $v_u=\mathbb E_{\mathcal D}[\kappa_u d_u]$ be the mean descent direction.
Independent pair-level context sampling gives $v_q=v_0+q\Delta v$, where $\Delta v=v_c-v_0$ is the rubric's extra update direction.

Fix the evaluation prompt law $\nu$, binary success rule $o_B$, and factor scores $z=(z_1,\ldots,z_8)^\top$.
Assume finite prompt and response spaces and a positive, continuously differentiable policy.
Rubric-free deployment uses $P_\theta(\vx,\vy)=\nu(\vx)\pi_\theta(\vy\mid\vx,0)$, with success probability $p=p_B(\theta)=\mathbb E_{P_\theta}[o_B]\in(0,1)$.
The extra direction changes deployment log-probabilities at rate
\begin{equation}
 h(\vx,\vy)=\left\langle\nabla_\theta\log\pi_\theta(\vy\mid\vx,0),\Delta v\right\rangle.
 \label{eq:prob_transport}
\end{equation}
Let $\bm D=\mathbb E[z\mid o_B=1]-\mathbb E[z\mid o_B=0]$ be population demand under $P_\theta$, and $a$ the least-squares coefficients of $h$ on centered factor scores.
Write $\epsilon_h,\epsilon_o$ for the root-mean-square projection residuals of $h$ and $o_B-p$, respectively; the full construction is in \appref{sec:prob_success_proof}.

\begin{proposition}[Local success from semantic alignment]
\label{prop:prob_success_odds}
At a shared parameter state and reference, with the update directions held fixed,
\begin{equation}
 \left\langle\nabla_\theta p_B,v_q-v_0\right\rangle
 =q\,p(1-p)\bigl(\bm D^\top a+\mathcal R\bigr),
 \label{eq:prob_success_alignment}
\end{equation}
where $|\mathcal R|\le\epsilon_o\epsilon_h/[p(1-p)]$.
For $q>0$, if $\bm D^\top a>\epsilon_o\epsilon_h/[p(1-p)]$, then
$p_B(\theta+\eta v_q)>p_B(\theta+\eta v_0)$ for all sufficiently small $\eta>0$.
The comparison reverses when $\bm D^\top a<-\epsilon_o\epsilon_h/[p(1-p)]$.
\end{proposition}

Demand $\bm D$ describes the behaviors associated with success; $a$ expresses the rubric-induced change in deployment log-probabilities in those semantic coordinates.
Their alignment sets the sign of the local success effect when it dominates the residual contribution outside the measured factors.
Changing the preference corpus can change the update directions and hence this alignment, even under the same rubric.

For two rendered rubrics $c_0,c_1$ at a shared policy and reference, replacing $\Delta v$ by $v_{c_1}-v_{c_0}$ gives the local edit comparison at a common exposure rate.
This comparison motivates the matched \textsc{keep}/\textsc{refine} continuation in \cref{sec:recursive}, which measures the edit's incremental effect over further training with the existing rubric.

\section{Experiments}
\label{sec:whatchanges}

\vspace{-3mm}
The analysis in \cref{sec:transfer} organizes three empirical questions: how rubric text changes the training update, how conditioning affects rubric-free outcomes across datasets, and what an edit adds beyond continuing with the existing rubric.
We examine them through fixed-state probes and a criterion-identity pilot, dataset-matched ordinary-DPO controls, and matched \textsc{keep}/\textsc{refine} continuations, respectively.

\paragraph{Models and Datasets.}
We cover two families and three sizes: Qwen3-4B-Instruct and Qwen3-8B-Instruct from the Qwen3 family~\citep{yang2025qwen3}, and Mistral-7B-Instruct-v0.3 from the Mistral family~\citep{jiang2023mistral}.
Formal runs train a rank-$64$ LoRA adapter ($\alpha=128$, dropout $0.05$) at learning rate $10^{-6}$, global batch $64$, DPO $\beta=1$. 
We use two public preference datasets, HelpSteer3~\citep{wang2025helpsteer3} and Tulu3~\citep{lambert2025tulu3}.

\paragraph{Benchmark Classes.}
Evaluation covers two benchmark classes, summarized separately.
The \emph{comprehensive-capability suite} is seven accuracy benchmarks---GSM8K~\citep{cobbe2021verifiers}, MATH-Hard~\citep{hendrycks2021math}, BBH~\citep{suzgun2022bbh}, MuSR~\citep{sprague2024musr}, ARC~\citep{clark2018arc}, TruthfulQA~\citep{lin2022truthfulqa}, and MMLU-Pro~\citep{wang2024mmlupro}---each with a task check defining the binary success outcome $o_B$ in \cref{prop:prob_success_odds}.
The \emph{open-ended generative suite} is three protocols---MT-Bench~\citep{zheng2023judging}, Arena-Hard~\citep{li2024arenahard}, and WildBench~\citep{lin2024wildbench}---on which a judge scores free-form responses and there is no task check.
Generative protocols are scored on matched item sets under one judge.

\paragraph{Feedback and Evaluation.}
\label{sec:feedback_evaluation}
The refinement engine reads item-level outputs from first-round rubric-free benchmark evaluations: judge rationales on Arena-Hard, MT-Bench, and WildBench, and correctness feedback on GSM8K, MATH-Hard, and MMLU-Pro.
The \textsc{keep} and \textsc{refine} continuations are then compared on matched evaluation items under the same benchmark protocols.
This evaluates a rubric update informed by benchmark feedback, with the unchanged-rubric branch controlling for additional training.
Feedback items overlap the final evaluation set, so the continuation comparison measures adaptation to the observed benchmark feedback.
The diagnostic/held-out split in \cref{sec:diagnosis} assesses measurement stability.
Generation, judging, extraction, paired tests, and all absolute scores are in \appref{sec:setup_details}.

\subsection{How Rubric Text Changes Learning}
\label{sec:context_readouts}
\paragraph{Context-Dependent Updates.}
In \cref{eq:decomposition}, context enters the update through both the margin-dependent weight $\kappa_u$ and the pair direction $d_u$.
A fixed-state probe over $7{,}253$ preference pairs gives conditioned--unconditioned gradient cosines of $0.74$--$0.79$ and relative gradient changes of $0.65$--$0.72$.
The likelihood probe measures the semantic sensitivity of the pairwise margin: the $f_5$ completeness criterion's margin-change slope is $0.0226$ nats per response token per unit of $\Delta z_5$, compared with $0.0116$ when read against reasoning contrast $\Delta z_2$.

\paragraph{Criterion Identity at Fixed Exposure.}
\label{sec:criterion_identity}
Changing the criterion text can change $v_c$ in \cref{sec:transfer} while preserving the conditioning interface.
A $150$-step pilot compares reasoning ($f_2$) and clarity ($f_7$) criteria of nearly equal length, using the same pairs and conditioned rows at $q=0.2$.
MuSR pass@4 is $70.8\%$ under the reasoning criterion, $68.6\%$ under the clarity criterion, and $70.8\%$ under ordinary DPO.
The paired $f_2$--$f_7$ comparison has $17$ prompts passing only under $f_2$ and $6$ only under $f_7$ (\appref{sec:pilot}).
The two criterion texts produce different rubric-free outcomes on MuSR pass@4, with the reasoning criterion matching ordinary DPO there.

\subsection{Dataset Contrasts and Rubric-Free Outcomes}
\begin{figure}[!htbp]
\centering
\includegraphics[width=\linewidth]{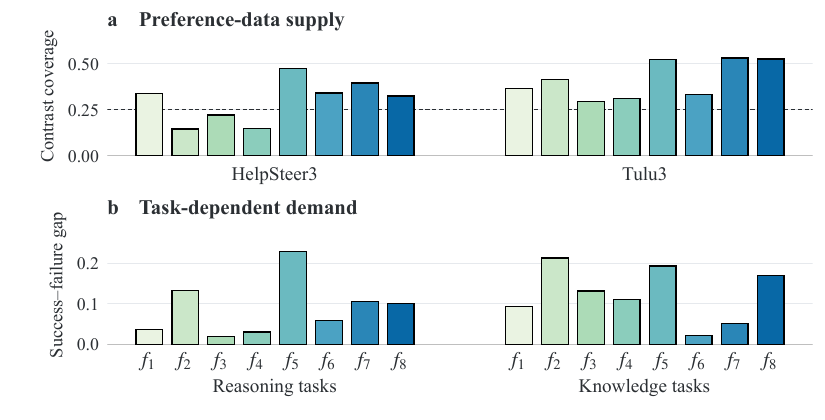}
\vspace{-7mm}
\caption{Diagnosis on shared semantic factors. (a) Contrast coverage (fraction of pairs) on both corpora using the same frozen judge; the dashed line marks the $0.25$ sparsity threshold. (b) Qwen3-4B demand, averaged equally over four reasoning or three knowledge benchmarks. Demand is the success--failure gap on the $0$--$4$ judge scale. Factors follow \appref{sec:ontology}.}
\label{fig:diagnosis}
\end{figure}

\paragraph{The Two Datasets Contrast Different Factors.}
The shared coordinates of \cref{sec:setup_notation} are what make the two datasets comparable.
\Cref{fig:diagnosis}a reports contrast coverage $C_k$ on both, and \cref{tab:ev_supply} gives the HelpSteer3 statistics behind it.
HelpSteer3 leaves three factors sparse against the $0.25$ sparsity rule: $f_2$ reasoning validity ($0.146$), $f_4$ factual and evidential grounding ($0.149$), and $f_3$ verification and calibration ($0.220$).
Reasoning validity ($f_2$) is meaningful on five of the seven benchmarks but is contrasted in only $14.6\%$ of HelpSteer3 pairs, while completeness ($f_5$) is meaningful on four and contrasted in $47.4\%$.
Tulu3 contrasts seven of the eight factors more often than HelpSteer3 does, and no factor of its eight falls below that rule; its lowest coverage is $0.294$.
Its largest advantages over HelpSteer3 are on $f_2$ ($+0.268$), $f_8$ ($+0.199$), $f_4$ ($+0.161$), and $f_7$ ($+0.135$).
These profiles describe the frequency of behavioral contrasts on the same axes used for task demand in \cref{sec:diagnosis}.
The full demand matrix is \cref{tab:demand_detail}, and across diagnostic and held-out subsets its largest change is $0.045$, with every gap keeping its sign.
\Cref{fig:diagnosis}b groups the same gaps by task family: verification and calibration separates success from failure about six times more strongly on the knowledge benchmarks than on the reasoning benchmarks ($0.131$ against $0.020$), and clarity and organization runs the other way ($0.051$ against $0.105$).

\begin{figure}[!htbp]
\centering
\includegraphics[width=\linewidth]{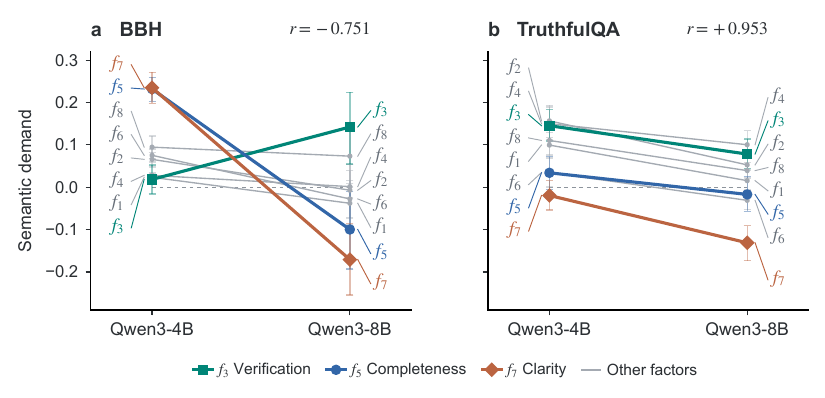}
\vspace{-7mm}
\caption{Semantic demand in Qwen3-4B and Qwen3-8B. Lines connect the same factor; error bars show recorded $95\%$ confidence intervals. $r$ is Pearson correlation across the eight factors.}
\label{fig:model_demand_slopes}
\end{figure}

Demand profiles differ between the two Qwen3 models in a task-dependent way (\cref{fig:model_demand_slopes}).
On BBH, the eight-factor profiles of Qwen3-4B and Qwen3-8B are negatively correlated ($r=-0.751$): completeness and clarity change from positive to negative gaps, while verification and calibration becomes more strongly associated with success.
TruthfulQA retains a similar relative profile across the two models ($r=0.953$), although the gap magnitudes change.
These measurements support diagnosing semantic priorities for each model and task.

\vspace{-3mm}

\vspace{-3mm}

\vspace{-3mm}
\paragraph{Conditioning Effects Across Datasets.}
\label{sec:openended}
\Cref{tab:main_results} reports the Tulu3 capability results and the HelpSteer3 open-ended results for both Qwen3 models.
BBH improves in both Qwen3 models, with effects measured against their own ordinary-DPO controls.
On Qwen3-4B, conditioning raises BBH from $15.43$ to $19.64$ ($+4.20$) and MATH-Hard from $39.15$ to $43.35$, and the seven-task average from $48.20$ to $50.02$.
The other five benchmarks move by at most $2.28$ points.
Qwen3-8B gains $9.61$ BBH points.
The Mistral-7B Tulu3 run decreases BBH by $7.95$ points relative to ordinary DPO (\cref{tab:mistral_results}; \appref{sec:ev_mistral}).

On Qwen3-4B, the four rubric compositions raise Arena-Hard by $1.4$--$2.2$ percentage points relative to ordinary DPO (\cref{tab:ablation_detail}). Common$+$Specific has the highest WildBench point estimate, while the full rubric has the highest MT-Bench score and average relative change across the three protocols.

\vspace{-3mm}
\begin{table}[!htbp]
\centering
\tablestyle
\caption{Rubric-free Qwen3 results on both benchmark suites, each from its own dataset. Capability scores and their unweighted seven-task mean are percentages from the Tulu3 runs. The open-ended protocols come from the HelpSteer3 runs: MT-Bench and WildBench use native scores and Arena-Hard is the win rate against GPT-4-0314 (\%). Open-ended $\mathrm{Avg.}\,\Delta=\frac{100}{3}\sum_b(s_b/s_b^{\mathrm{DPO}}-1)$ averages relative changes across all three benchmarks against the same block's DPO control. Bold marks maxima within each model block (including ties). Shading marks rubric-trained rows.}
\label{tab:main_results}
\setlength{\tabcolsep}{1.6pt}
\fontsize{8}{9.5}\selectfont
\begin{tabular}{lrrrrrrrrrrrr}
\toprule
 & \multicolumn{8}{c}{\shortstack{\textbf{Tulu3}\\\textbf{comprehensive capability (\%)}}} & \multicolumn{4}{c}{\shortstack{\textbf{HelpSteer3}\\\textbf{open-ended generation}}}\\
\cmidrule(lr){2-9}\cmidrule(lr){10-13}
\textbf{Method} & \textbf{GSM8K} & \textbf{\shortstack{MATH-\\Hard}} & \textbf{BBH} & \textbf{MuSR} & \textbf{ARC} & \textbf{\shortstack{Truthful\\QA}} & \textbf{\shortstack{MMLU-\\Pro}} & \textbf{Avg.} & \textbf{\shortstack{MT-\\Bench}} & \textbf{\shortstack{Arena-\\Hard (\%)}} & \textbf{\shortstack{Wild-\\Bench}} & \textbf{\shortstack{Avg.\ $\Delta$\\(\%)}}\\
\midrule
\tablepanel{13}{Qwen3-4B}
Instruct & $85.90$ & $\mathbf{62.24}$ & $12.82$ & $\mathbf{43.39}$ & $\mathbf{62.80}$ & $52.27$ & $41.98$ & $\mathbf{51.63}$ & $6.320$ & $49.0$ & $\mathbf{3.730}$ & $-2.96$\\
Ordinary DPO & $84.53$ & $39.15$ & $15.43$ & $42.59$ & $61.09$ & $52.71$ & $41.88$ & $48.20$ & $6.890$ & $50.9$ & $3.617$ & $0.00$\\
\rowcolor{tablerubric}
\Pone & $\mathbf{86.81}$ & $43.35$ & $19.64$ & $43.12$ & $61.35$ & $53.74$ & $\mathbf{42.10}$ & $50.02$ & $\mathbf{6.930}$ & $\mathbf{53.0}$ & $3.727$ & $\mathbf{+2.58}$\\
\rowcolor{tablerubric}
\Ptwo & $80.74$ & $37.12$ & $\mathbf{22.73}$ & $42.06$ & $59.47$ & $\mathbf{55.16}$ & $40.43$ & $48.24$ & \multicolumn{4}{c}{---}\\
\midrule
\tablepanel{13}{Qwen3-8B}
Instruct & $\mathbf{91.90}$ & $70.13$ & $37.43$ & $54.90$ & $94.10$ & $\mathbf{78.20}$ & $67.60$ & $70.61$ & $7.410$ & $64.5$ & $4.074$ & $-0.05$\\
Ordinary DPO & $91.80$ & --- & $38.66$ & $\mathbf{55.20}$ & $94.65$ & $76.80$ & $67.35$ & --- & $7.219$ & $64.2$ & $4.211$ & $0.00$\\
\rowcolor{tablerubric}
\Pone & $91.40$ & $\mathbf{72.77}$ & $\mathbf{48.26}$ & $55.15$ & $\mathbf{95.00}$ & $77.85$ & $\mathbf{68.30}$ & $\mathbf{72.68}$ & $\mathbf{7.416}$ & $\mathbf{65.7}$ & $\mathbf{4.329}$ & $\mathbf{+2.62}$\\
\bottomrule
\end{tabular}
\end{table}

\begin{samepage}
\paragraph{Variation Across Datasets and Models.}
\label{sec:dataset_effects}
In \cref{prop:prob_success_odds}, the local effect depends on task demand and the corpus-dependent update direction.
The ordinary-DPO controls measure the full rubric's finite-training effect separately within each dataset.
On Qwen3-4B, the capability effect is larger on Tulu3: BBH improves by $4.20$ points, compared with $0.23$ on HelpSteer3, and the seven-task mean by $1.82$ points, compared with $0.18$ (\cref{tab:main_results,tab:ablation_detail}).
Both methods train for one epoch within each dataset.
On HelpSteer3, Qwen3-8B gains $0.20$ MT-Bench points and $1.5$ Arena-Hard points (\cref{tab:main_results}).
Mistral-7B gains $0.29$ MT-Bench points and $1.8$ Arena-Hard points (\cref{tab:mistral_results}).
\par
\end{samepage}

\begin{table}[!htbp]
\centering
\tablestyle
\caption{Mistral-7B results on both benchmark suites, each from its own dataset. Capability scores and their unweighted seven-task mean are percentages. Open-ended Avg.\ $\Delta$ follows the  definition in \cref{tab:main_results}. Bold marks maxima within each dataset block; shading marks rubric-trained rows.}
\label{tab:mistral_results}
\setlength{\tabcolsep}{1.6pt}
\fontsize{8}{9.5}\selectfont
\begin{tabular}{lrrrrrrrrrrrr}
\toprule
 & \multicolumn{8}{c}{\shortstack{\textbf{Tulu3}\\\textbf{comprehensive capability (\%)}}} & \multicolumn{4}{c}{\shortstack{\textbf{HelpSteer3}\\\textbf{open-ended generation}}}\\
\cmidrule(lr){2-9}\cmidrule(lr){10-13}
\textbf{Method} & \textbf{GSM8K} & \textbf{\shortstack{MATH-\\Hard}} & \textbf{BBH} & \textbf{MuSR} & \textbf{ARC} & \textbf{\shortstack{Truthful\\QA}} & \textbf{\shortstack{MMLU-\\Pro}} & \textbf{Avg.} & \textbf{\shortstack{MT-\\Bench}} & \textbf{\shortstack{Arena-\\Hard (\%)}} & \textbf{\shortstack{Wild-\\Bench}} & \textbf{\shortstack{Avg.\ $\Delta$\\(\%)}}\\
\midrule
Instruct & $45.75$ & $19.65$ & $36.95$ & $\mathbf{52.45}$ & $\mathbf{79.45}$ & $\mathbf{58.75}$ & $30.05$ & $\mathbf{46.15}$ & $5.090$ & $18.1$ & $2.336$ & $-8.44$\\
Ordinary DPO & $44.45$ & $20.90$ & $\mathbf{40.40}$ & $50.95$ & $76.50$ & $53.90$ & $30.00$ & $45.30$ & $4.850$ & $24.0$ & $2.477$ & $0.00$\\
\rowcolor{tablerubric}
\Pone & $\mathbf{47.90}$ & $\mathbf{21.50}$ & $32.45$ & $51.85$ & $75.60$ & $55.40$ & $\mathbf{32.15}$ & $45.26$ & $\mathbf{5.140}$ & $\mathbf{25.8}$ & $\mathbf{2.480}$ & $\mathbf{+4.53}$\\
\bottomrule
\end{tabular}
\end{table}

\paragraph{Adding Factors to a Common Rubric.}
\label{sec:ablation}
Four HelpSteer3 compositions each contain the Common factors $f_2$ and $f_5$, with the Specific or Low-demand factors added.
Their seven-task means differ by at most $0.13$ points and exceed ordinary DPO's $52.56$ by at most $0.18$.
Common raises BBH by $0.90$ points relative to ordinary DPO, adding Specific yields $0.47$, adding Low yields $0.40$, and including both yields $0.23$.
All four compositions improve Arena-Hard by $1.4$--$2.2$ points, with a different ordering from BBH.
The small average differences among compositions do not identify a superior factor set.
The full ontology supplies a fixed set of clauses for subsequent editing.
Per-benchmark results are in \cref{tab:ablation_detail}.

\paragraph{Rubric Dependence at Inference Time.}
\label{sec:weights}
Reinserting a rubric at inference time has task-dependent effects on the same trained checkpoint (\appref{sec:reinsertion_protocol}).
On matched sampled BBH items, the trained $R_1$ wording changes accuracy by $-1.25$ points and the candidate $R_2$ by $+3.70$ points.
The separate greedy MATH-Hard evaluation changes by $+0.68$ and $+0.76$ points respectively.

\vspace{-3mm}
\subsection{What the Rubric Edit Contributes}
\label{sec:recursive}

\vspace{-3mm}

The edit comparison in \cref{sec:transfer} fixes the starting policy and varies only the rubric-induced update direction; it is carried through a matched continuation, with \textsc{keep} controlling for additional training. The trained Qwen3-4B policy reviews its first-round feedback and finds a recurring $f_4$ gap in its training rubric.
The frozen clause prohibited unsupported assertions but left the evidence requirement for ``common knowledge'' ambiguous.
The revised Prefer/Avoid pair reads:
\begin{quote}\ttfamily\small\raggedright
\textcolor{criterionfavor}{Prefer:} factual claims are grounded in the given information or \textbf{explicitly sourced} common knowledge.\\
\textcolor{criterionavoid}{Avoid:} invented facts, unsupported assertions, or \textbf{inferences from vague common knowledge}.
\end{quote}
Round 1 refines only $f_4$; the other seven factors return \textsc{keep} because no recurring uncovered requirement is identified.
The edit adds seven words across the Prefer/Avoid pair in a $234$-token rubric.
\Cref{tab:continuation_main} evaluates it through two HelpSteer3 continuations from the same Qwen3-4B policy, using the same pairs, frozen reference, optimizer settings, seed, and pair-level exposure mask.
The \textsc{keep} branch retains $R_0$; the \textsc{refine} branch uses $R_1$.
Unchanged-rubric continuation raises BBH from $14.96$ to $18.05$, a gain of $3.09$ points.
Relative to that continuation, the edit changes BBH by $-0.25$ points, MT-Bench by $+0.406$, and Arena-Hard by $-0.1$ percentage points.
The edit's incremental effects differ across tasks: BBH decreases slightly, while the MT-Bench point estimate recovers part of the decline under unchanged-rubric continuation, from $6.93$ in round 1 to $6.419$ under \textsc{keep} and $6.825$ under \textsc{refine}.
Evaluation protocols and paired comparisons are in \appref{sec:paired}.

\vspace{-3mm}
\begin{table}[ht]
\centering
\tablestyle
\caption{Qwen3-4B matched continuation on HelpSteer3. Both branches start from \Pone{} and differ only in rubric text. Evaluation is rubric-free. Open-ended Avg.\ $\Delta$ follows the relative-DPO definition in \cref{tab:main_results}. Bold marks column maxima.}
\label{tab:continuation_main}
\footnotesize
\setlength{\tabcolsep}{3pt}
\begin{tabular}{lrrrrrc}
\toprule
\textbf{Method} & \textbf{BBH} & \textbf{MATH-Hard} & \textbf{MT-Bench} & \textbf{Arena-Hard} & \textbf{WildBench} & \textbf{\shortstack{Avg.\ $\Delta$ (\%)}}\\
\midrule
\Pone{} & $14.96$ & $\mathbf{64.18}$ & $\mathbf{6.93}$ & $53.0$ & $3.727$ & $+2.58$\\
\PtwoK{} & $\mathbf{18.05}$ & $63.82$ & $6.419$ & $\mathbf{54.6}$ & $3.781$ & $+1.66$\\
\rowcolor{tablerubric}
\PtwoR{} & $17.80$ & $63.72$ & $6.825$ & $54.5$ & $\mathbf{3.836}$ & $\mathbf{+4.06}$\\
\bottomrule
\end{tabular}
\end{table}

The revision trace records how failure attribution and edit constraints determine the next rubric.
Factor $f_3$ receives the strongest failure signal, but its existing Avoid clause already names the concrete requirement.
The engine nevertheless proposes an $f_3$ rewrite.
The gate rejects it because it drops part of the frozen semantic anchor; cross-factor vocabulary also triggers a warning.
Factor $f_3$ therefore returns to \textsc{keep}.
This rejection preserves the original $f_3$ clause for further learning.
On Qwen3-8B, attribution marks every requirement in the feedback set as already covered and the operator returns \textsc{keep} for all factors.
The subsequent 4B feedback pass produces a three-factor $R_2$ candidate for a future training round.

\section{Conclusion}
\vspace{-3mm}

RubricOpt connects policy self-evaluation to revision of the behavioral specification guiding preference learning.
Its semantic rubric translates failure feedback into training-time criteria while preference pairs, labels, and the optimization objective remain fixed.
Experiments across comprehensive capability and open-ended generation show that rubric semantics can shape rubric-free behavior, with benefits depending on the model and preference corpus.
The tested revision yields mixed incremental results over unchanged-rubric continuation.
This makes the behavioral specification itself adaptable through the policy's feedback loop.

\vspace{-3mm}
\paragraph{Scope and Future Work.}
The study uses an associative demand measure and a manually specified ontology.
Evaluating successive revisions on feedback-disjoint tasks and comparing proposals from initial and updated policies under a shared training budget would test the scope for sustained self-improvement.

\section*{Reproducibility Statement}
Every trained method in this paper is a rank-$64$ LoRA adapter over a released checkpoint, and the appendices specify the dataset, optimizer, schedule, seed, conditioning rate, reference convention, evaluation protocols, and judge configuration for every reported run.
The two preference datasets are public, and the seven accuracy benchmarks and three open-ended protocols are used through their released forms.
The initial formal rubrics can be reconstructed from the canonical bank in \appref{sec:ontology} and the template in \appref{sec:prompts}.
The accepted HelpSteer3 $R_1$ edit is quoted in \cref{sec:recursive}.
The refinement operator is specified as five stages with their inputs, outputs, and the deterministic acceptance checks (\appref{sec:workflow_full}).

\section*{Broader Impact Statement}
RubricOpt exposes behavioral criteria as short, inspectable training artifacts, with an evidence trail for each accepted edit.
The policy under diagnosis proposes the edits, so its interpretation of failures can reinforce its own biases.
Deterministic checks enforce evidence references and textual constraints; rubric-free evaluation measures the resulting behavior.
Deployments should review proposed rubric changes against their intended behaviors and evaluate each edit with a matched continuation control.

\bibliography{iclr2027_conference}

\begin{thebibliography}{49}
\providecommand{\natexlab}[1]{#1}
\providecommand{\url}[1]{\texttt{#1}}
\expandafter\ifx\csname urlstyle\endcsname\relax
  \providecommand{\doi}[1]{doi: #1}\else
  \providecommand{\doi}{doi: \begingroup \urlstyle{rm}\Url}\fi

\bibitem[Ashury-Tahan et~al.(2026)Ashury-Tahan, Mai, Bandel, Shmueli-Scheuer,
  and Choshen]{ashury2026erroratlas}
Shir Ashury-Tahan, Yifan Mai, Elron Bandel, Michal Shmueli-Scheuer, and Leshem
  Choshen.
\newblock {ErrorMap} and {ErrorAtlas}: Charting the failure landscape of large
  language models.
\newblock \emph{arXiv preprint arXiv:2601.15812}, 2026.
\newblock \doi{10.48550/arXiv.2601.15812}.
\newblock URL \url{https://arxiv.org/abs/2601.15812}.

\bibitem[Askell et~al.(2021)Askell, Bai, Chen, Drain, Ganguli, Henighan, Jones,
  Joseph, Mann, DasSarma, Elhage, Hatfield-Dodds, Hernandez, Kernion, Ndousse,
  Olsson, Amodei, Brown, Clark, McCandlish, Olah, and
  Kaplan]{askell2021general}
Amanda Askell, Yuntao Bai, Anna Chen, Dawn Drain, Deep Ganguli, Tom Henighan,
  Andy Jones, Nicholas Joseph, Ben Mann, Nova DasSarma, Nelson Elhage, Zac
  Hatfield-Dodds, Danny Hernandez, Jackson Kernion, Kamal Ndousse, Catherine
  Olsson, Dario Amodei, Tom Brown, Jack Clark, Sam McCandlish, Chris Olah, and
  Jared Kaplan.
\newblock A general language assistant as a laboratory for alignment.
\newblock \emph{arXiv preprint arXiv:2112.00861}, 2021.
\newblock \doi{10.48550/arXiv.2112.00861}.
\newblock URL \url{https://arxiv.org/abs/2112.00861}.

\bibitem[Azar et~al.(2024)Azar, Guo, Piot, Munos, Rowland, Valko, and
  Calandriello]{azar2024ipo}
Mohammad~Gheshlaghi Azar, Zhaohan~Daniel Guo, Bilal Piot, R{\'e}mi Munos, Mark
  Rowland, Michal Valko, and Daniele Calandriello.
\newblock A general theoretical paradigm to understand learning from human
  preferences.
\newblock In \emph{Proceedings of the 27th International Conference on
  Artificial Intelligence and Statistics}, volume 238 of \emph{Proceedings of
  Machine Learning Research}, pp.\  4447--4455. PMLR, 2024.
\newblock URL \url{https://proceedings.mlr.press/v238/gheshlaghi-azar24a.html}.

\bibitem[Bai et~al.(2022)Bai, Jones, Ndousse, Askell, Chen, DasSarma, Drain,
  Fort, Ganguli, Henighan, Joseph, Kadavath, Kernion, Conerly, El-Showk,
  Elhage, Hatfield-Dodds, Hernandez, Hume, Johnston, Kravec, Lovitt, Nanda,
  Olsson, Amodei, Brown, Clark, McCandlish, Olah, Mann, and
  Kaplan]{bai2022harmless}
Yuntao Bai, Andy Jones, Kamal Ndousse, Amanda Askell, Anna Chen, Nova DasSarma,
  Dawn Drain, Stanislav Fort, Deep Ganguli, Tom Henighan, Nicholas Joseph,
  Saurav Kadavath, Jackson Kernion, Tom Conerly, Sheer El-Showk, Nelson Elhage,
  Zac Hatfield-Dodds, Danny Hernandez, Tristan Hume, Scott Johnston, Shauna
  Kravec, Liane Lovitt, Neel Nanda, Catherine Olsson, Dario Amodei, Tom Brown,
  Jack Clark, Sam McCandlish, Chris Olah, Ben Mann, and Jared Kaplan.
\newblock Training a helpful and harmless assistant with reinforcement learning
  from human feedback, 2022.
\newblock URL \url{https://arxiv.org/abs/2204.05862}.

\bibitem[Chen et~al.(2025)Chen, Li, Li, Yang, Paisley, and
  Zeng]{chen2025dequantified}
Wei Chen, Shigui Li, Jiacheng Li, Junmei Yang, John Paisley, and Delu Zeng.
\newblock Dequantified diffusion-schr\"odinger bridge for density ratio
  estimation.
\newblock In \emph{International Conference on Machine Learning}, 2025.
\newblock URL \url{https://openreview.net/forum?id=zvyHCOcwsw}.

\bibitem[Chen et~al.(2026)Chen, Wu, Yang, Zeng, Zhao, Paisley, Chen, and
  Wang]{chen2026towards}
Wei Chen, Yubing Wu, Junmei Yang, Delu Zeng, Qibin Zhao, John Paisley, Min
  Chen, and Zhou Wang.
\newblock Towards disentangled preference optimization dynamics: Suppress the
  loser, preserve the winner.
\newblock In \emph{International Conference on Machine Learning}, 2026.
\newblock URL \url{https://openreview.net/forum?id=TaNH4XiQ6P}.

\bibitem[Clark et~al.(2018)Clark, Cowhey, Etzioni, Khot, Sabharwal, Schoenick,
  and Tafjord]{clark2018arc}
Peter Clark, Isaac Cowhey, Oren Etzioni, Tushar Khot, Ashish Sabharwal, Carissa
  Schoenick, and Oyvind Tafjord.
\newblock Think you have solved question answering? {Try} {ARC}, the {AI2}
  reasoning challenge, 2018.
\newblock URL \url{https://arxiv.org/abs/1803.05457}.

\bibitem[Cobbe et~al.(2021)Cobbe, Kosaraju, Bavarian, Chen, Jun, Kaiser,
  Plappert, Tworek, Hilton, Nakano, Hesse, and Schulman]{cobbe2021verifiers}
Karl Cobbe, Vineet Kosaraju, Mohammad Bavarian, Mark Chen, Heewoo Jun, Lukasz
  Kaiser, Matthias Plappert, Jerry Tworek, Jacob Hilton, Reiichiro Nakano,
  Christopher Hesse, and John Schulman.
\newblock Training verifiers to solve math word problems, 2021.
\newblock URL \url{https://arxiv.org/abs/2110.14168}.

\bibitem[Ding et~al.(2026)Ding, Huang, Fang, Liao, Li, Zhang, Wu, Zhao, and
  Wang]{ding2026evorubrics}
Hongxin Ding, Baixiang Huang, Yue Fang, Weibin Liao, Zheng Li, Jinyang Zhang,
  Zhijing Wu, Junfeng Zhao, and Yasha Wang.
\newblock {EvoRubrics}: Dynamic rubrics as rewards via adversarial co-evolution
  for {LLM} reinforcement learning, 2026.
\newblock URL \url{https://arxiv.org/abs/2606.23038}.

\bibitem[Ethayarajh et~al.(2024)Ethayarajh, Xu, Muennighoff, Jurafsky, and
  Kiela]{ethayarajh2024kto}
Kawin Ethayarajh, Winnie Xu, Niklas Muennighoff, Dan Jurafsky, and Douwe Kiela.
\newblock Model alignment as prospect theoretic optimization.
\newblock In \emph{Proceedings of the 41st International Conference on Machine
  Learning}, volume 235 of \emph{Proceedings of Machine Learning Research},
  pp.\  12634--12651. PMLR, 2024.
\newblock URL \url{https://proceedings.mlr.press/v235/ethayarajh24a.html}.

\bibitem[Gallego(2025)]{gallego2025cpt}
V{\'i}ctor Gallego.
\newblock Configurable preference tuning with rubric-guided synthetic data.
\newblock \emph{arXiv preprint arXiv:2506.11702}, 2025.
\newblock \doi{10.48550/arXiv.2506.11702}.
\newblock URL \url{https://arxiv.org/abs/2506.11702}.
\newblock ICML 2025 Workshop on Models of Human Feedback for AI Alignment.

\bibitem[Gunjal et~al.(2025)Gunjal, Wang, Lau, Nath, He, Liu, and
  Hendryx]{gunjal2025rar}
Anisha Gunjal, Anthony Wang, Elaine Lau, Vaskar Nath, Yunzhong He, Bing Liu,
  and Sean Hendryx.
\newblock Rubrics as rewards: Reinforcement learning beyond verifiable domains.
\newblock \emph{arXiv preprint arXiv:2507.17746}, 2025.
\newblock \doi{10.48550/arXiv.2507.17746}.
\newblock URL \url{https://arxiv.org/abs/2507.17746}.

\bibitem[Hendrycks et~al.(2021)Hendrycks, Burns, Kadavath, Arora, Basart, Tang,
  Song, and Steinhardt]{hendrycks2021math}
Dan Hendrycks, Collin Burns, Saurav Kadavath, Akul Arora, Steven Basart, Eric
  Tang, Dawn Song, and Jacob Steinhardt.
\newblock Measuring mathematical problem solving with the {MATH} dataset, 2021.
\newblock URL \url{https://arxiv.org/abs/2103.03874}.

\bibitem[Ho \& Salimans(2022)Ho and Salimans]{ho2022cfg}
Jonathan Ho and Tim Salimans.
\newblock Classifier-free diffusion guidance.
\newblock \emph{arXiv preprint arXiv:2207.12598}, 2022.
\newblock \doi{10.48550/arXiv.2207.12598}.
\newblock URL \url{https://arxiv.org/abs/2207.12598}.
\newblock NeurIPS 2021 Workshop on Deep Generative Models and Downstream
  Applications.

\bibitem[Jiang et~al.(2023)Jiang, Sablayrolles, Mensch, Bamford, Chaplot,
  de~las Casas, Bressand, Lengyel, Lample, Saulnier, Lavaud, Lachaux, Stock,
  Scao, Lavril, Wang, Lacroix, and Sayed]{jiang2023mistral}
Albert~Q. Jiang, Alexandre Sablayrolles, Arthur Mensch, Chris Bamford,
  Devendra~Singh Chaplot, Diego de~las Casas, Florian Bressand, Gianna Lengyel,
  Guillaume Lample, Lucile Saulnier, Lélio~Renard Lavaud, Marie-Anne Lachaux,
  Pierre Stock, Teven~Le Scao, Thibaut Lavril, Thomas Wang, Timothée Lacroix,
  and William~El Sayed.
\newblock Mistral 7b, 2023.
\newblock URL \url{https://arxiv.org/abs/2310.06825}.

\bibitem[Lambert et~al.(2025)Lambert, Morrison, Pyatkin, Huang, Ivison,
  Brahman, Miranda, Liu, Dziri, Lyu, Gu, Malik, Graf, Hwang, Yang, Le~Bras,
  Tafjord, Wilhelm, Soldaini, Smith, Wang, Dasigi, and
  Hajishirzi]{lambert2025tulu3}
Nathan Lambert, Jacob Morrison, Valentina Pyatkin, Shengyi Huang, Hamish
  Ivison, Faeze Brahman, Lester James~V. Miranda, Alisa Liu, Nouha Dziri, Shane
  Lyu, Yuling Gu, Saumya Malik, Victoria Graf, Jena~D. Hwang, Jiangjiang Yang,
  Ronan Le~Bras, Oyvind Tafjord, Chris Wilhelm, Luca Soldaini, Noah~A. Smith,
  Yizhong Wang, Pradeep Dasigi, and Hannaneh Hajishirzi.
\newblock Tulu 3: Pushing frontiers in open language model post-training, 2025.
\newblock URL \url{https://arxiv.org/abs/2411.15124}.

\bibitem[Li et~al.(2024)Li, Chiang, Frick, Dunlap, Wu, Zhu, Gonzalez, and
  Stoica]{li2024arenahard}
Tianle Li, Wei-Lin Chiang, Evan Frick, Lisa Dunlap, Tianhao Wu, Banghua Zhu,
  Joseph~E. Gonzalez, and Ion Stoica.
\newblock From crowdsourced data to high-quality benchmarks: Arena-hard and
  benchbuilder pipeline, 2024.
\newblock URL \url{https://arxiv.org/abs/2406.11939}.

\bibitem[Lin et~al.(2024)Lin, Deng, Chandu, Brahman, Ravichander, Pyatkin,
  Dziri, Le~Bras, and Choi]{lin2024wildbench}
Bill~Yuchen Lin, Yuntian Deng, Khyathi Chandu, Faeze Brahman, Abhilasha
  Ravichander, Valentina Pyatkin, Nouha Dziri, Ronan Le~Bras, and Yejin Choi.
\newblock {WildBench}: Benchmarking {LLMs} with challenging tasks from real
  users in the wild, 2024.
\newblock URL \url{https://arxiv.org/abs/2406.04770}.

\bibitem[Lin et~al.(2022)Lin, Hilton, and Evans]{lin2022truthfulqa}
Stephanie Lin, Jacob Hilton, and Owain Evans.
\newblock {TruthfulQA}: Measuring how models mimic human falsehoods, 2022.
\newblock URL \url{https://arxiv.org/abs/2109.07958}.

\bibitem[Liu et~al.(2025)Liu, Bai, Lu, Sun, Kong, Wang, Shan, Jose, Liu, Wen,
  Yu, and Cao]{liu2024tisdpo}
Aiwei Liu, Haoping Bai, Zhiyun Lu, Yanchao Sun, Xiang Kong, Xiaoming Wang,
  Jiulong Shan, Albin~Madappally Jose, Xiaojiang Liu, Lijie Wen, Philip~S. Yu,
  and Meng Cao.
\newblock {TIS-DPO}: Token-level importance sampling for direct preference
  optimization with estimated weights.
\newblock In \emph{The Thirteenth International Conference on Learning
  Representations}, 2025.
\newblock URL
  \url{https://proceedings.iclr.cc/paper_files/paper/2025/hash/7fb9f39075a5202472676a7531568212-Abstract-Conference.html}.

\bibitem[Madaan et~al.(2023)Madaan, Tandon, Gupta, Hallinan, Gao, Wiegreffe,
  Alon, Dziri, Prabhumoye, Yang, Gupta, Majumder, Hermann, Welleck,
  Yazdanbakhsh, and Clark]{madaan2023selfrefine}
Aman Madaan, Niket Tandon, Prakhar Gupta, Skyler Hallinan, Luyu Gao, Sarah
  Wiegreffe, Uri Alon, Nouha Dziri, Shrimai Prabhumoye, Yiming Yang, Shashank
  Gupta, Bodhisattwa~Prasad Majumder, Katherine Hermann, Sean Welleck, Amir
  Yazdanbakhsh, and Peter Clark.
\newblock Self-refine: Iterative refinement with self-feedback.
\newblock In \emph{Advances in Neural Information Processing Systems},
  volume~36, 2023.
\newblock \doi{10.52202/075280-2019}.
\newblock URL
  \url{https://papers.neurips.cc/paper_files/paper/2023/hash/91edff07232fb1b55a505a9e9f6c0ff3-Abstract-Conference.html}.

\bibitem[Meng et~al.(2024)Meng, Xia, and Chen]{meng2024simpo}
Yu~Meng, Mengzhou Xia, and Danqi Chen.
\newblock {SimPO}: Simple preference optimization with a reference-free reward.
\newblock In \emph{Advances in Neural Information Processing Systems},
  volume~37, 2024.
\newblock \doi{10.52202/079017-3946}.
\newblock URL
  \url{https://proceedings.neurips.cc/paper_files/paper/2024/hash/e099c1c9699814af0be873a175361713-Abstract-Conference.html}.

\bibitem[Ouyang et~al.(2022)Ouyang, Wu, Jiang, Almeida, Wainwright, Mishkin,
  Zhang, Agarwal, Slama, Ray, Schulman, Hilton, Kelton, Miller, Simens, Askell,
  Welinder, Christiano, Leike, and Lowe]{ouyang2022training}
Long Ouyang, Jeff Wu, Xu~Jiang, Diogo Almeida, Carroll~L. Wainwright, Pamela
  Mishkin, Chong Zhang, Sandhini Agarwal, Katarina Slama, Alex Ray, John
  Schulman, Jacob Hilton, Fraser Kelton, Luke Miller, Maddie Simens, Amanda
  Askell, Peter Welinder, Paul Christiano, Jan Leike, and Ryan Lowe.
\newblock Training language models to follow instructions with human feedback.
\newblock In \emph{Advances in Neural Information Processing Systems},
  volume~35, 2022.
\newblock URL \url{https://arxiv.org/abs/2203.02155}.

\bibitem[Pryzant et~al.(2023)Pryzant, Iter, Li, Lee, Zhu, and
  Zeng]{pryzant2023automatic}
Reid Pryzant, Dan Iter, Jerry Li, Yin Lee, Chenguang Zhu, and Michael Zeng.
\newblock Automatic prompt optimization with ``gradient descent'' and beam
  search.
\newblock In Houda Bouamor, Juan Pino, and Kalika Bali (eds.),
  \emph{Proceedings of the 2023 Conference on Empirical Methods in Natural
  Language Processing}, pp.\  7957--7968, Singapore, December 2023. Association
  for Computational Linguistics.
\newblock \doi{10.18653/v1/2023.emnlp-main.494}.
\newblock URL \url{https://aclanthology.org/2023.emnlp-main.494/}.

\bibitem[Rafailov et~al.(2023)Rafailov, Sharma, Mitchell, Manning, Ermon, and
  Finn]{rafailov2023dpo}
Rafael Rafailov, Archit Sharma, Eric Mitchell, Christopher~D. Manning, Stefano
  Ermon, and Chelsea Finn.
\newblock Direct preference optimization: Your language model is secretly a
  reward model.
\newblock In \emph{Advances in Neural Information Processing Systems},
  volume~36, 2023.
\newblock \doi{10.52202/075280-2338}.
\newblock URL
  \url{https://proceedings.neurips.cc/paper_files/paper/2023/hash/a85b405ed65c6477a4fe8302b5e06ce7-Abstract-Conference.html}.

\bibitem[Snell et~al.(2022)Snell, Klein, and Zhong]{snell2022context}
Charlie Snell, Dan Klein, and Ruiqi Zhong.
\newblock Learning by distilling context.
\newblock \emph{arXiv preprint arXiv:2209.15189}, 2022.
\newblock \doi{10.48550/arXiv.2209.15189}.
\newblock URL \url{https://arxiv.org/abs/2209.15189}.

\bibitem[Sprague et~al.(2024)Sprague, Ye, Bostrom, Chaudhuri, and
  Durrett]{sprague2024musr}
Zayne Sprague, Xi~Ye, Kaj Bostrom, Swarat Chaudhuri, and Greg Durrett.
\newblock {MuSR}: Testing the limits of chain-of-thought with multistep soft
  reasoning, 2024.
\newblock URL \url{https://arxiv.org/abs/2310.16049}.

\bibitem[Stiennon et~al.(2020)Stiennon, Ouyang, Wu, Ziegler, Lowe, Voss,
  Radford, Amodei, and Christiano]{stiennon2020learning}
Nisan Stiennon, Long Ouyang, Jeff Wu, Daniel~M. Ziegler, Ryan Lowe, Chelsea
  Voss, Alec Radford, Dario Amodei, and Paul Christiano.
\newblock Learning to summarize from human feedback.
\newblock In \emph{Advances in Neural Information Processing Systems},
  volume~33, 2020.
\newblock URL \url{https://arxiv.org/abs/2009.01325}.

\bibitem[Sugiyama et~al.(2012)Sugiyama, Suzuki, and
  Kanamori]{sugiyama2012density}
Masashi Sugiyama, Taiji Suzuki, and Takafumi Kanamori.
\newblock Density-ratio matching under the bregman divergence: a unified
  framework of density-ratio estimation.
\newblock \emph{Annals of the Institute of Statistical Mathematics},
  64\penalty0 (5):\penalty0 1009--1044, 2012.

\bibitem[Suzgun et~al.(2023)Suzgun, Scales, Sch{\"a}rli, Gehrmann, Tay, Chung,
  Chowdhery, Le, Chi, Zhou, and Wei]{suzgun2022bbh}
Mirac Suzgun, Nathan Scales, Nathanael Sch{\"a}rli, Sebastian Gehrmann, Yi~Tay,
  Hyung~Won Chung, Aakanksha Chowdhery, Quoc~V. Le, Ed~H. Chi, Denny Zhou, and
  Jason Wei.
\newblock Challenging {BIG-Bench} tasks and whether chain-of-thought can solve
  them.
\newblock In \emph{The Eleventh International Conference on Learning
  Representations}, 2023.
\newblock URL \url{https://arxiv.org/abs/2210.09261}.
\newblock arXiv:2210.09261.

\bibitem[Vapnik \& Vashist(2009)Vapnik and Vashist]{vapnik2009lupi}
Vladimir Vapnik and Akshay Vashist.
\newblock A new learning paradigm: Learning using privileged information.
\newblock \emph{Neural Networks}, 22\penalty0 (5--6):\penalty0 544--557, 2009.
\newblock \doi{10.1016/j.neunet.2009.06.042}.
\newblock URL \url{https://doi.org/10.1016/j.neunet.2009.06.042}.

\bibitem[Wang et~al.(2026{\natexlab{a}})Wang, Zheng, Liu, Cao, Zhang, Chen,
  Yang, and Chen]{wang2026serpo}
Jianze Wang, Kunwang Zheng, Ying Liu, Yu~Cao, Qilong Zhang, Jinlong Chen, Hua
  Yang, and Qianglong Chen.
\newblock {SERPO}: Self-evolving rubric policy optimization for open-ended
  test-time reinforcement learning, 2026{\natexlab{a}}.
\newblock URL \url{https://arxiv.org/abs/2607.26873}.

\bibitem[Wang et~al.(2026{\natexlab{b}})Wang, Li, Xu, Zhou, Shi, Cai, Zhang,
  Li, and Zhang]{wang2026apter}
Xukai Wang, Liangqi Li, Zhiyue Xu, Jingang Zhou, Xiaoyu Shi, Jiansheng Cai,
  Bo~Zhang, Zhe Li, and Xu-Yao Zhang.
\newblock {APTER}: Adaptive post-training with expert-grounded rubrics,
  2026{\natexlab{b}}.
\newblock URL \url{https://arxiv.org/abs/2608.14212}.

\bibitem[Wang et~al.(2024)Wang, Ma, Zhang, Ni, Chandra, Guo, Ren, Arulraj, He,
  Jiang, Li, Ku, Wang, Zhuang, Fan, Yue, and Chen]{wang2024mmlupro}
Yubo Wang, Xueguang Ma, Ge~Zhang, Yuansheng Ni, Abhranil Chandra, Shiguang Guo,
  Weiming Ren, Aaran Arulraj, Xuan He, Ziyan Jiang, Tianle Li, Max Ku, Kai
  Wang, Alex Zhuang, Rongqi Fan, Xiang Yue, and Wenhu Chen.
\newblock {MMLU-Pro}: A more robust and challenging multi-task language
  understanding benchmark, 2024.
\newblock URL \url{https://arxiv.org/abs/2406.01574}.

\bibitem[Wang et~al.(2025)Wang, Zeng, Delalleau, Shin, Soares, Bukharin, Evans,
  Dong, and Kuchaiev]{wang2025helpsteer3}
Zhilin Wang, Jiaqi Zeng, Olivier Delalleau, Hoo-Chang Shin, Felipe Soares,
  Alexander Bukharin, Ellie Evans, Yi~Dong, and Oleksii Kuchaiev.
\newblock Helpsteer3-preference: Open human-annotated preference data across
  diverse tasks and languages, 2025.
\newblock URL \url{https://arxiv.org/abs/2505.11475}.

\bibitem[Wei et~al.(2026)Wei, Ding, Li, Ma, Weng, Cai, Ding, and
  Zhang]{wei2026diagevo}
Xincheng Wei, Yifan Ding, Yoshua Li, Dongsheng Ma, Rongxiang Weng, Xunliang
  Cai, Wenjian Ding, and Yao Zhang.
\newblock {DiagEvo}: Diagnosis-guided self-evolution via hierarchical error
  memory, 2026.
\newblock URL \url{https://arxiv.org/abs/2609.00768v1}.

\bibitem[Wu et~al.(2025{\natexlab{a}})Wu, Yuan, Golovneva, Xu, Tian, Jiao,
  Weston, and Sukhbaatar]{wu2025metarewarding}
Tianhao Wu, Weizhe Yuan, Olga Golovneva, Jing Xu, Yuandong Tian, Jiantao Jiao,
  Jason~E. Weston, and Sainbayar Sukhbaatar.
\newblock Meta-rewarding language models: Self-improving alignment with
  {LLM}-as-a-meta-judge.
\newblock In \emph{Proceedings of the 2025 Conference on Empirical Methods in
  Natural Language Processing}, Suzhou, China, 2025{\natexlab{a}}. Association
  for Computational Linguistics.
\newblock \doi{10.18653/v1/2025.emnlp-main.583}.
\newblock URL \url{https://aclanthology.org/2025.emnlp-main.583/}.

\bibitem[Wu et~al.(2025{\natexlab{b}})Wu, Sun, Yuan, Ji, Yang, and
  Gu]{wu2025sppo}
Yue Wu, Zhiqing Sun, Huizhuo Yuan, Kaixuan Ji, Yiming Yang, and Quanquan Gu.
\newblock Self-play preference optimization for language model alignment.
\newblock In \emph{The Thirteenth International Conference on Learning
  Representations}, 2025{\natexlab{b}}.
\newblock URL \url{https://arxiv.org/abs/2405.00675}.
\newblock arXiv:2405.00675.

\bibitem[Xiao et~al.(2025)Xiao, Yuan, Li, Chen, and Honavar]{xiao2025dil}
Teng Xiao, Yige Yuan, Mingxiao Li, Zhengyu Chen, and Vasant~G. Honavar.
\newblock On a connection between imitation learning and {RLHF}.
\newblock In \emph{The Thirteenth International Conference on Learning
  Representations}, 2025.
\newblock URL \url{https://arxiv.org/abs/2503.05079}.
\newblock arXiv:2503.05079.

\bibitem[Yang et~al.(2025{\natexlab{a}})Yang, Li, Yang, Zhang, Hui, Zheng, Yu,
  Gao, Huang, Lv, Zheng, Liu, Zhou, Huang, Hu, Ge, Wei, Lin, Tang, Yang, Tu,
  Zhang, Yang, Yang, Zhou, Zhou, Lin, Dang, Bao, Yang, Yu, Deng, Li, Xue, Li,
  Zhang, Wang, Zhu, Men, Gao, Liu, Luo, Li, Tang, Yin, Ren, Wang, Zhang, Ren,
  Fan, Su, Zhang, Zhang, Wan, Liu, Wang, Cui, Zhang, Zhou, and
  Qiu]{yang2025qwen3}
An~Yang, Anfeng Li, Baosong Yang, Beichen Zhang, Binyuan Hui, Bo~Zheng, Bowen
  Yu, Chang Gao, Chengen Huang, Chenxu Lv, Chujie Zheng, Dayiheng Liu, Fan
  Zhou, Fei Huang, Feng Hu, Hao Ge, Haoran Wei, Huan Lin, Jialong Tang, Jian
  Yang, Jianhong Tu, Jianwei Zhang, Jianxin Yang, Jiaxi Yang, Jing Zhou,
  Jingren Zhou, Junyang Lin, Kai Dang, Keqin Bao, Kexin Yang, Le~Yu, Lianghao
  Deng, Mei Li, Mingfeng Xue, Mingze Li, Pei Zhang, Peng Wang, Qin Zhu, Rui
  Men, Ruize Gao, Shixuan Liu, Shuang Luo, Tianhao Li, Tianyi Tang, Wenbiao
  Yin, Xingzhang Ren, Xinyu Wang, Xinyu Zhang, Xuancheng Ren, Yang Fan, Yang
  Su, Yichang Zhang, Yinger Zhang, Yu~Wan, Yuqiong Liu, Zekun Wang, Zeyu Cui,
  Zhenru Zhang, Zhipeng Zhou, and Zihan Qiu.
\newblock Qwen3 technical report, 2025{\natexlab{a}}.
\newblock URL \url{https://arxiv.org/abs/2505.09388}.

\bibitem[Yang et~al.(2025{\natexlab{b}})Yang, Liu, Xie, Huang, Min, and
  Ananiadou]{yang2024sepo}
Kailai Yang, Zhiwei Liu, Qianqian Xie, Jimin Huang, Erxue Min, and Sophia
  Ananiadou.
\newblock Selective preference optimization via token-level reward function
  estimation.
\newblock In \emph{Proceedings of the 2025 Conference on Empirical Methods in
  Natural Language Processing}, pp.\  7032--7056, Suzhou, China,
  2025{\natexlab{b}}. Association for Computational Linguistics.
\newblock \doi{10.18653/v1/2025.emnlp-main.359}.
\newblock URL \url{https://aclanthology.org/2025.emnlp-main.359/}.

\bibitem[Yang et~al.(2026)Yang, Lin, Liu, Tian, Liu, and Zhang]{yang2025tidpo}
Ning Yang, Hai Lin, Yibo Liu, Baoliang Tian, Guoqing Liu, and Haijun Zhang.
\newblock Token-importance guided direct preference optimization.
\newblock In \emph{The Fourteenth International Conference on Learning
  Representations}, 2026.
\newblock URL
  \url{https://proceedings.iclr.cc/paper_files/paper/2026/hash/172fe9f8cc55953bad5c24774bf0142b-Abstract-Conference.html}.
\newblock ICLR 2026 Poster.

\bibitem[Ye et~al.(2024)Ye, Kim, Kim, Hwang, Kim, Jo, Thorne, Kim, and
  Seo]{ye2023flask}
Seonghyeon Ye, Doyoung Kim, Sungdong Kim, Hyeonbin Hwang, Seungone Kim, Yongrae
  Jo, James Thorne, Juho Kim, and Minjoon Seo.
\newblock {FLASK}: Fine-grained language model evaluation based on alignment
  skill sets.
\newblock In \emph{The Twelfth International Conference on Learning
  Representations}, 2024.
\newblock URL \url{https://arxiv.org/abs/2307.10928}.
\newblock ICLR 2024 Spotlight; arXiv:2307.10928.

\bibitem[Yuan et~al.(2024)Yuan, Pang, Cho, Li, Sukhbaatar, Xu, and
  Weston]{yuan2024selfrewarding}
Weizhe Yuan, Richard~Yuanzhe Pang, Kyunghyun Cho, Xian Li, Sainbayar
  Sukhbaatar, Jing Xu, and Jason~E. Weston.
\newblock Self-rewarding language models.
\newblock In \emph{Proceedings of the 41st International Conference on Machine
  Learning}, volume 235 of \emph{Proceedings of Machine Learning Research},
  pp.\  57905--57923. PMLR, 2024.
\newblock URL \url{https://proceedings.mlr.press/v235/yuan24d.html}.

\bibitem[Zeng et~al.(2024)Zeng, Liu, Ma, Yang, Zhang, and Wang]{zeng2024tdpo}
Yongcheng Zeng, Guoqing Liu, Weiyu Ma, Ning Yang, Haifeng Zhang, and Jun Wang.
\newblock Token-level direct preference optimization.
\newblock In \emph{Proceedings of the 41st International Conference on Machine
  Learning}, volume 235 of \emph{Proceedings of Machine Learning Research},
  pp.\  58348--58365. PMLR, 2024.
\newblock URL \url{https://proceedings.mlr.press/v235/zeng24c.html}.

\bibitem[Zeng et~al.(2025)Zeng, Wang, Hajishirzi, and Koh]{zeng2025evaltree}
Zhiyuan Zeng, Yizhong Wang, Hannaneh Hajishirzi, and Pang~Wei Koh.
\newblock {EvalTree}: Profiling language model weaknesses via hierarchical
  capability trees.
\newblock In \emph{Second Conference on Language Modeling}, 2025.
\newblock URL \url{https://arxiv.org/abs/2503.08893}.
\newblock arXiv:2503.08893.

\bibitem[Zhang et~al.(2026)Zhang, He, Wang, Lv, Ma, Lin, Zhong, and
  Guo]{zhang2026cort}
Bo-Wen Zhang, Junwei He, Wen Wang, Song-Lin Lv, Wentao Ma, Rongyi Lin, Shuhan
  Zhong, and Lan-Zhe Guo.
\newblock {CoRT}: Counterfactual replay for token-level rubric-guided policy
  optimization.
\newblock \emph{arXiv preprint arXiv:2607.25659}, 2026.
\newblock \doi{10.48550/arXiv.2607.25659}.
\newblock URL \url{https://arxiv.org/abs/2607.25659}.

\bibitem[Zheng et~al.(2023)Zheng, Chiang, Sheng, Zhuang, Wu, Zhuang, Lin, Li,
  Li, Xing, Zhang, Gonzalez, and Stoica]{zheng2023judging}
Lianmin Zheng, Wei-Lin Chiang, Ying Sheng, Siyuan Zhuang, Zhanghao Wu, Yonghao
  Zhuang, Zi~Lin, Zhuohan Li, Dacheng Li, Eric Xing, Hao Zhang, Joseph
  Gonzalez, and Ion Stoica.
\newblock Judging llm-as-a-judge with mt-bench and chatbot arena.
\newblock In A.~Oh, T.~Naumann, A.~Globerson, K.~Saenko, M.~Hardt, and
  S.~Levine (eds.), \emph{Advances in Neural Information Processing Systems},
  volume~36, pp.\  46595--46623. Curran Associates, Inc., 2023.
\newblock \doi{10.52202/075280-2020}.
\newblock URL
  \url{https://proceedings.neurips.cc/paper_files/paper/2023/file/91f18a1287b398d378ef22505bf41832-Paper-Datasets_and_Benchmarks.pdf}.

\bibitem[Zhou et~al.(2025)Zhou, Zhang, Zhao, and Meng]{zhou2025treg}
Wenxuan Zhou, Shujian Zhang, Lingxiao Zhao, and Tao Meng.
\newblock {T-REG}: Preference optimization with token-level reward
  regularization.
\newblock In \emph{Proceedings of the 63rd Annual Meeting of the Association
  for Computational Linguistics (Volume 1: Long Papers)}, pp.\  27876--27889,
  Vienna, Austria, 2025. Association for Computational Linguistics.
\newblock \doi{10.18653/v1/2025.acl-long.1353}.
\newblock URL \url{https://aclanthology.org/2025.acl-long.1353/}.

\end{thebibliography}
\bibliographystyle{iclr2027_conference}
\clearpage
\appendix

\paragraph{LLM Usage Disclosure.}
Large language models (LLMs) were used throughout this work to assist with several aspects of the research and writing process.
Specifically, we used LLMs to improve the clarity and presentation of the manuscript, to support literature retrieval and the discovery of relevant prior work, and to assist in drafting portions of the paper.
The authors reviewed, revised, and verified all LLM-assisted content and remain fully responsible for the final manuscript.

\appendixcontentstrue
\begingroup
\renewcommand{\contentsname}{Appendix Contents}
\setcounter{tocdepth}{2}
\hypersetup{linkcolor=black}
\tableofcontents
\endgroup

\section{Preliminaries and Measurement}

\subsection{Notation}
\label{sec:notation}
\paragraph{Models and data} $m,\pi_\theta,\pi_{\mathrm{ref}}$: Current model, its trainable policy, and the frozen reference.
$B,o$: Diagnostic task and its binary outcome.
$\mathcal D,\vx,\vy,\vy^+,\vy^-$: Preference data, prompt, response, and chosen/rejected responses.
$L$: Number of response tokens in a response; $\bar\ell$ averages a log-probability over them.

\paragraph{Behavioral measurements} $\mathcal F=\{f_1,\ldots,f_8\}$: Fixed factor ontology of eight factors.
$z_k(\vx,\vy)\in\{0,\ldots,4\}$: Ordinal score a blind judge assigns to factor $f_k$.
$D_{m,B,k}$: Model- and task-specific semantic demand (\cref{eq:demand}).
$\Delta z_k,\mu_k,C_k,P_k$: Pair contrast, its mean, contrast coverage, and label-agreement rate (\cref{eq:supply}).

\paragraph{Conditioning and updates} $\mathcal G\subseteq\mathcal F,c_r,\emptycrit$: Selected factors (all eight in the main method), the context rendered from rubric $R_r$ by \cref{eq:rubric_context}, and the absent criterion.
$u,u_t,q(\rho_t)$: Conditioning context $u\in\{c_r,\emptycrit\}$, its draw at training step $t$ within round $r$, and the conditioning rate at progress $\rho_t$.
$\beta,\sigma,\mathcal L$: DPO scale, logistic function, and Semantic Rubric Steering loss.
$M_u,s_u,g_u,\kappa_u$: Length-normalized chosen--rejected margin, mean per-token response score gradient, loss gradient, and scalar multiplier, all in context $u$.
$\nu,A_k,H_k$: Evaluation prompt distribution, rubric-free expected factor score $\mathbb E_{\nu}[\bar z_k(\vx)]$, and its gradient Lipschitz constant.
$\bm{G}_c,\bm{G}_0,\bm{G}_q$: Mean loss gradients with and without the criterion, and their mixture at rate $q$.
$\eta,\theta_q,\theta_0,r_k$: Step size, two local updated parameter states, and the transfer remainder.
$\bar z_k(\vx),\bm{\delta}_c,\bm{\delta}_0$: Prompt-specific expected factor score and two actual optimizer increments from a shared state.
$\mathrm{Shift}_k$: Semantic shift, the change in mean factor score after an update.

\paragraph{Rounds and probabilistic analysis} $R_r,\pi_{r+1},\mathcal B_{r+1}$: Round-$r$ rubric, the policy trained under its context $c_r$, and the resulting behavioral feedback (\cref{eq:rubricopt}).
$\psi_0,F$: Full-response deployment score (\cref{eq:deployment_score}) and its Fisher matrix (\cref{eq:prob_fisher_length}).
$\bm D,\Sigma,\Gamma_u,\mathcal P_{\tau:T}$: Deployment demand vector and factor covariance (\cref{eq:prob_demand_covariance}), training--deployment gradient alignment (\cref{eq:transport_detail}), and remaining Fisher path length.
$\PtwoK$: Matched re-optimization branch that keeps the criterion $R_0$.
$\PtwoR$: Matched re-optimization branch that refines the criterion to $R_1$. The two branches differ only in the criterion.

\subsection{Measurement and the Factor Ontology}
\label{sec:ontology}
The measurements use the definitions of \cref{sec:setup_notation} and the blind judge of \appref{sec:judge}.

\paragraph{Worked Demand Example.}
A GSM8K problem asks the cost of six pounds of butter at \$4 per pound when every second pound is half price.
A successful response answers \$18, while a failed response answers \$16.
Their judge vectors are $[2,3,1,3,2,3,3,3]$ and $[2,1,1,3,2,3,3,3]$, differing only on reasoning validity.
This illustrates scoring, not estimation of demand, which aggregates the success--failure gap over the benchmark population.

\paragraph{Worked Supply Example.}
In one HelpSteer3 joke-request pair, the chosen--rejected score difference is $[0,0,0,0,+1,0,0,0]$: it contrasts completeness but not reasoning validity.
A pair about female US presidents contrasts several factors at once, with difference $[+2,+2,0,+2,+1,+2,0,+1]$.
Supply therefore summarizes a distribution of within-pair contrasts rather than assigning each pair to one factor.

\paragraph{The Factor Ontology.}
Every quantity above is indexed by the fixed eight-factor ontology, whose demand profile over the seven benchmarks is \cref{tab:demand_detail}.
The frozen bank supplies one Prefer and one matched Avoid clause per factor, as rendered for training:
{\ttfamily\footnotesize\raggedright
\par\noindent
\textcolor{criterionhead}{f1 task comprehension}\\
\hspace*{1.4em}\textcolor{criterionfavor}{Prefer:} show a correct understanding of what was asked\\
\hspace*{1.4em}\textcolor{criterionavoid}{Avoid:} answering a different question or ignoring stated conditions\\

\textcolor{criterionhead}{f2 reasoning validity}\\
\hspace*{1.4em}\textcolor{criterionfavor}{Prefer:} reasoning steps that are logically grounded and internally\\
\hspace*{1.4em}consistent\\
\hspace*{1.4em}\textcolor{criterionavoid}{Avoid:} unsupported intermediate inferences or invalid deductions\\

\textcolor{criterionhead}{f3 verification and calibration}\\
\hspace*{1.4em}\textcolor{criterionfavor}{Prefer:} verify key intermediate results and express uncertainty in a\\
\hspace*{1.4em}calibrated way\\
\hspace*{1.4em}\textcolor{criterionavoid}{Avoid:} unchecked steps or unsupported overconfidence\\

\textcolor{criterionhead}{f4 factual and evidential grounding}\\
\hspace*{1.4em}\textcolor{criterionfavor}{Prefer:} factual claims are grounded in the given information or common\\
\hspace*{1.4em}knowledge\\
\hspace*{1.4em}\textcolor{criterionavoid}{Avoid:} invented facts or unsupported assertions\\

\textcolor{criterionhead}{f5 completeness}\\
\hspace*{1.4em}\textcolor{criterionfavor}{Prefer:} responses that include the reasoning and information needed to\\
\hspace*{1.4em}fully address the task\\
\hspace*{1.4em}\textcolor{criterionavoid}{Avoid:} missing necessary reasoning, information, or important parts of\\
\hspace*{1.4em}the task\\

\textcolor{criterionhead}{f6 instruction adherence}\\
\hspace*{1.4em}\textcolor{criterionfavor}{Prefer:} follow the stated instructions and stay on task\\
\hspace*{1.4em}\textcolor{criterionavoid}{Avoid:} drifting off-topic or violating requested formats\\

\textcolor{criterionhead}{f7 clarity and organization}\\
\hspace*{1.4em}\textcolor{criterionfavor}{Prefer:} clearly expressed and logically organized\\
\hspace*{1.4em}\textcolor{criterionavoid}{Avoid:} jumbled steps or answers that are hard to locate\\

\textcolor{criterionhead}{f8 efficiency and conciseness}\\
\hspace*{1.4em}\textcolor{criterionfavor}{Prefer:} keep only the necessary content\\
\hspace*{1.4em}\textcolor{criterionavoid}{Avoid:} repeating reasoning or padding with filler
}

\subsection{Judge Configuration and Reliability}
\label{sec:judge}
\paragraph{Production Configuration.}
Semantic factor scores reported in this paper come from a self-judge. The policy under diagnosis scores its own rubric-free outputs with one fixed, version-tagged judge prompt.
Decoding is greedy (temperature $0$, top-$p$ $1$, top-$k$ $1$) with at most $1100$ new tokens, at batch size $64$.
The self-judge is blind to the task outcome, the gold answer, the preference role, and the identity of the generating model.
Each call returns one raw integer score per factor together with a supporting sentence, and the scores are not normalized across factors, responses, or runs.

\paragraph{Score Anchors.}
The shared scale is $0$ = clearly poor or absent, $1$ = weak, $2$ = mixed or partial, $3$ = good, and $4$ = strong.
The production prompt defines the top of the scale as an absolute standard rather than a comparative one: a $4$ is earned by concrete extra rigor, such as re-checking a computation, stating and handling an assumption or edge case, noticing an ambiguity, or flagging its own uncertainty, and it is explicitly not awarded for being correct and fluent.
A $3$ is competent, complete, routine work with no extra rigor; a $2$ does the minimum with visible gaps; a $1$ is missing or merely asserts an important behavior.
The prompt also asks the judge to list two strengths and two weaknesses before scoring, and gives per-factor rules that cap a score at $1$.
For reasoning validity, the cap applies when any step is a non-sequitur or an equation does not follow.
The prompt permits \texttt{NA} only when a behavior cannot occur for that response kind, and states that a required behavior which is missing still scores $0$ or $1$ rather than \texttt{NA}.
In practice that rule never fired: across the $108{,}685$ scored responses, every factor was assessable on every response.

\paragraph{Agreement Audits.}
A different judge model rescored $1{,}200$ responses.
On items where the audit judge gave a decisive score, mean agreement with the production self-judge was $0.741$.
Per factor, agreement was highest for task comprehension ($0.909$), factual and evidential grounding ($0.857$), and instruction adherence ($0.800$), and lowest for clarity and organization ($0.633$), verification and calibration ($0.616$), and efficiency and conciseness ($0.559$).
\Cref{tab:ev_judge} gives the per-factor agreement.
The three generative benchmarks are scored by MiniMax-M3.

\begin{table}[!htbp]
\centering
\tablestyle
\caption{Blind pairwise agreement across four benchmarks. Pairwise verdicts are compared with the direction of independent per-response scores. Decisive excludes ties in either judgment; agreement rates use this subset.}
\label{tab:ev_judge}
\footnotesize
\begin{tabular}{lrrr}
\toprule
\textbf{Factor} & \textbf{Decisive} & \textbf{Agree} & \textbf{Decisive-only rate}\\
\midrule
$f_1$ task comprehension & $11$ & $10$ & $0.9091$\\
$f_2$ reasoning validity & $52$ & $41$ & $0.7885$\\
$f_3$ verification and calibration & $232$ & $143$ & $0.6164$\\
$f_4$ factual and evidential grounding & $7$ & $6$ & $0.8571$\\
$f_5$ completeness & $105$ & $80$ & $0.7619$\\
$f_6$ instruction adherence & $15$ & $12$ & $0.8000$\\
$f_7$ clarity and organization & $199$ & $126$ & $0.6332$\\
$f_8$ efficiency and conciseness & $247$ & $138$ & $0.5587$\\
\midrule
\rowcolor{tablegroup}
\textbf{Mean} & & & $\mathbf{0.7406}$\\
\bottomrule
\end{tabular}
\end{table}

\paragraph{Factor Separability.}
Across $108{,}685$ complete responses, the production self-judge yields mean absolute inter-factor correlation $0.195$ and maximum $0.362$.
The older judge prompt, used as a collapse control, reaches maximum correlation $1.000$ and mean $0.378$, and produces only four distinct score vectors across $288$ responses, whereas the production prompt produces $822$.
The eight factors are therefore not collapsed onto one quality axis in this analysis.

\section{Proofs}

\subsection{Local Behavioral Transfer}
\label{sec:transfer_proof}
\paragraph{The Deployment Space.}
The analysis uses finite prompt and response spaces and a strictly positive, continuously differentiable policy on a neighborhood of every path considered.
The full-response deployment score is
\begin{equation}
 \psi_0(\vx,\vy)=\nabla_\theta\log\pi_\theta(\vy\mid\vx,0)
 =L(\vy)s_0(\vy),
 \label{eq:deployment_score}
\end{equation}
For a training triple $\xi=(\vx',\vy^+,\vy^-)\sim\mathcal D$, write $d_u(\xi)$ for its pair direction.
The incremental rate of change of deployment log-probability along the rubric-induced direction is
\begin{align}
 \Gamma_u(\vx,\vy;\xi)&=\langle \psi_0(\vx,\vy),d_u(\xi)\rangle,\nonumber\\
 h(\vx,\vy)&=\langle \psi_0(\vx,\vy),\Delta v\rangle
 =\mathbb E_{\xi\sim\mathcal D}
   [\kappa_c \Gamma_c(\vx,\vy;\xi)
    -\kappa_0 \Gamma_0(\vx,\vy;\xi)].
 \label{eq:transport_detail}
\end{align}
The score identity gives $\mathbb E_{P_\theta}[\psi_0\mid\vx]=0$ and hence $\mathbb E_{P_\theta}[h\mid\vx]=0$, so each fixed preference pair contributes through the alignment of its training gradient with the deployment score.

\paragraph{The Behavioral Readout.}
What remains is to characterize that direction.
For a rubric-free evaluation distribution $\nu$, define the behavioural readout
\begin{equation}
A_k(\theta)=\mathbb E_{\vx\sim\nu,\,\vy\sim\pi_\theta(\cdot\mid \vx)}[z_k(\vx,\vy)],
\end{equation}
and let $\bm G_c$ and $\bm G_0$ be mean preference-loss gradients with and without the criterion, with $\bm G_q=(1-q)\bm G_0+q\bm G_c$.
\begin{proposition}
\label{prop:transfer}
Suppose $A_k$ has an $H_k$-Lipschitz gradient on a neighborhood containing the two update segments.
For $\theta_q=\theta-\eta\bm G_q$ and $\theta_0=\theta-\eta\bm G_0$,
\begin{align}
A_k(\theta_q)-A_k(\theta_0)
&=-\eta q\langle\nabla_\theta A_k,\bm G_c-\bm G_0\rangle+r_k,
\\
|r_k|&\le\frac{H_k\eta^2}{2}
\left(\|\bm G_q\|^2+\|\bm G_0\|^2\right).
\label{eq:transfer_remainder}
\end{align}
\end{proposition}

\paragraph{The Score-Function Identity.}
Assume differentiation and expectation can be interchanged and define $\bar z_k(\vx)=\mathbb E_{\vy\sim\pi_\theta(\cdot\mid \vx)}[z_k(\vx,\vy)]$ at the current parameter state.
The score-function identity gives
\begin{equation}
 \nabla_\theta A_k=
 \mathbb E_{\vx\sim\nu,\,\vy\sim\pi_\theta(\cdot\mid \vx)}
 \bigl[(z_k(\vx,\vy)-\bar z_k(\vx))\nabla_\theta\log\pi_\theta(\vy\mid \vx)\bigr].
 \label{eq:behavior_gradient}
\end{equation}
The baseline term vanishes because the conditional mean score gradient is zero.
A finite response space with a smooth, positive policy provides a sufficient setting for this identity, and the factor score itself need not be differentiable in the response: a randomized judge enters through its fixed conditional mean score.
\Cref{eq:behavior_gradient} is taken on generated responses under the rubric-free policy, which is a different quantity from the training-pair contrast $\Delta z_k$.

\paragraph{Actual Optimizer Steps.}
For two parameter increments $\bm{\delta}_c$ and $\bm{\delta}_0$ from the same parameter state, the same Taylor argument yields
\begin{equation}
 \begin{split}
 A_k(\theta+\bm{\delta}_c)-A_k(\theta+\bm{\delta}_0)
 &=\langle\nabla_\theta A_k,\bm{\delta}_c-\bm{\delta}_0\rangle+r_k,\\
 |r_k|&\le\tfrac12 H_k
 \bigl(\|\bm{\delta}_c\|^2+\|\bm{\delta}_0\|^2\bigr).
 \end{split}
 \label{eq:optimizer_transfer}
\end{equation}
Here the same smoothness assumption holds along both update segments, and the form accommodates adaptive updates and momentum, which is the regime of the experiments.

\Cref{eq:optimizer_transfer} compares updates from shared parameters, optimizer state, and minibatch.
Their effect on a behavior depends on the alignment of the increment difference with that behavior's gradient.
The context probes in \cref{sec:context_readouts} measure training-pair gradient and margin changes at a fixed state.
The benchmark comparisons measure rubric-free outcomes after training.

\paragraph{Proof Of \cref{prop:transfer}.}
For either $\bm{G}=\bm{G}_q$ or $\bm{G}=\bm{G}_0$, the Lipschitz-gradient assumption gives the Taylor expansion
\begin{equation}
 A_k(\theta-\eta\bm{G})=A_k(\theta)
 -\eta\langle\nabla_\theta A_k,\bm{G}\rangle+e(\bm{G}),
 \qquad |e(\bm{G})|\le\tfrac12 H_k\eta^2\|\bm{G}\|^2.
 \end{equation}
Subtract the two expansions and use $\bm{G}_q-\bm{G}_0=q(\bm{G}_c-\bm{G}_0)$.
The remainder is $r_k=e(\bm{G}_q)-e(\bm{G}_0)$, whose absolute value is bounded by \cref{eq:transfer_remainder} by the triangle inequality.
In particular, a sufficient condition for strict improvement over the ordinary step is
\begin{equation}
 -q\langle\nabla_\theta A_k,\bm{G}_c-\bm{G}_0\rangle
 >\frac{H_k\eta}{2}\bigl(\|\bm{G}_q\|^2+\|\bm{G}_0\|^2\bigr).
 \end{equation}
A nonzero first-order term dominates the quadratic remainder for sufficiently small positive $\eta$, proving the sign statements.
A zero inner product leaves the sign undetermined.
Improvement over the initial policy is a different comparison: its first-order term is $-\eta\langle\nabla_\theta A_k,\bm{G}_q\rangle$.\hfill$\square$

\subsection{Proof Of \texorpdfstring{\cref{prop:prob_success_odds}}{Prop.~1}}
\label{sec:prob_success_proof}
\paragraph{Semantic Projections and the Log-Odds Form.}
At a fixed state, write $p=p_B(\theta)$, $Z=z-\mathbb E[z]$, and
\begin{equation}
 \bm D=\mathbb E[z\mid o_B=1]-\mathbb E[z\mid o_B=0],\qquad
 \Sigma=\mathbb E[ZZ^\top],
 \label{eq:prob_demand_covariance}
\end{equation}
\begin{align}
 a&=\Sigma^\dagger\mathbb E[Zh],& r_h&=h-a^\top Z,
 &\epsilon_h^2&=\mathbb E[r_h^2],\nonumber\\
 b&=\Sigma^\dagger\mathbb E[Z(o_B-p)],& r_o&=o_B-p-b^\top Z,
 &\epsilon_o^2&=\mathbb E[r_o^2],
 \label{eq:prob_semantic_projection}
\end{align}
Here $\Sigma^\dagger$ is the Moore--Penrose pseudoinverse, $a$ is the coefficient vector for the log-probability rate $h$, and $\epsilon_h,\epsilon_o$ are the corresponding root-mean-square residuals.
The residual contribution is
\begin{equation}
 \mathcal R=\frac{\mathbb E[r_o r_h]}{p(1-p)},\qquad
 |\mathcal R|\le\frac{\epsilon_o\epsilon_h}{p(1-p)}.
 \label{eq:prob_residual_bound}
\end{equation}
If either projection residual vanishes, then $\mathcal R=0$.
For $\theta_q(\eta)=\theta+\eta v_q$ and $\theta_0(\eta)=\theta+\eta v_0$, with directions fixed at $\theta$ and $\operatorname{logit}(p)=\log[p/(1-p)]$, the equivalent log-odds form is
\begin{equation}
 \left.\frac{d}{d\eta}
 \left\{\operatorname{logit}p_B(\theta_q(\eta))
       -\operatorname{logit}p_B(\theta_0(\eta))\right\}\right|_{\eta=0}
 =q\bigl(\bm D^\top a+\mathcal R\bigr),
 \label{eq:prob_odds_identity}
\end{equation}

\paragraph{Proof.}
The deployment score satisfies $\mathbb E[\psi_0\mid\vx]=\sum_{\vy}\nabla_\theta\pi_\theta(\vy\mid\vx,0)=0$.
For any fixed direction $w$, differentiating the deployment expectation gives
\begin{equation}
 \nabla_\theta p_B^\top w
 =\mathbb E[(o_B-p)\langle\psi_0,w\rangle].
 \end{equation}
Taking $w=\Delta v$ and dividing by $p(1-p)$ expresses the incremental log-odds derivative as
\begin{equation}
 \frac{\mathbb E[(o_B-p)h]}{p(1-p)}
 =\mathbb E[h\mid o_B=1]-\mathbb E[h\mid o_B=0].
 \end{equation}
Likewise, conditioning on the binary outcome gives $\mathbb E[Z(o_B-p)]=p(1-p)\bm D$.

The normal equations for \cref{eq:prob_semantic_projection} hold even when $\Sigma$ is singular.
Indeed, $w\in\ker\Sigma$ implies $\mathbb E[(w^\top Z)^2]=0$, hence $w^\top Z=0$ almost surely.
Both covariance vectors being projected are therefore in the range of $\Sigma$.
Thus $\mathbb E[Zr_h]=\mathbb E[Zr_o]=0$, and
\begin{equation}
 \mathbb E[(o_B-p)h]
 =p(1-p)\bm D^\top a+\mathbb E[r_o r_h].
 \end{equation}
Cauchy--Schwarz bounds the last expectation by $\epsilon_o\epsilon_h$.
The least-squares residual variance for $o_B-p$ is $p(1-p)-\mathbb E[Z(o_B-p)]^\top\Sigma^\dagger\mathbb E[Z(o_B-p)]$, which gives \cref{eq:prob_residual_bound}.
Finally, $v_q-v_0=q\Delta v$ proves \cref{eq:prob_odds_identity}; multiplying by $p(1-p)$ gives \cref{eq:prob_success_alignment}.
A strictly positive derivative of the log-odds difference, which is zero at $\eta=0$, makes that difference positive for sufficiently small positive $\eta$.
The logit is strictly increasing.
The negative case follows identically.\hfill$\square$

\subsection{Retention After Rubric Withdrawal}
\label{sec:prob_retention_proof}
Let $\tau$ mark the end of rubric exposure and $T$ the final checkpoint, with success probability $p_t=p_B(\theta(t))$ and its monotone transform $\alpha_t=\arcsin\sqrt{p_t}$.
The Fisher path length $\mathcal P_{\tau:T}$ measures accumulated movement of the rubric-free response distribution over the remaining training.
\paragraph{Fisher Path Length and Probability Bounds.}
For the piecewise continuously differentiable path $\theta(t)$, define
\begin{equation}
 F(\theta)=\mathbb E_{P_\theta}[\psi_0\psi_0^\top],\qquad
 \mathcal P_{\tau:T}
 =\int_\tau^T
 \sqrt{\dot\theta(t)^\top F(\theta(t))\dot\theta(t)}\,dt.
 \label{eq:prob_fisher_length}
\end{equation}

\begin{proposition}[Retention after withdrawal]
\label{prop:prob_retention}
Along a piecewise continuously differentiable parameter path with $0<p_t<1$ and finite $\mathcal P_{\tau:T}$,
\begin{equation}
 |\alpha_T-\alpha_\tau|\le\tfrac12\mathcal P_{\tau:T},
 \label{eq:prob_retention_angle}
\end{equation}
For a fixed comparison success rate $p_\star\in(0,p_\tau)$, set $\alpha_\star=\arcsin\sqrt{p_\star}$.
If $\mathcal P_{\tau:T}<2(\alpha_\tau-\alpha_\star)$, then $p_T>p_\star$.
\end{proposition}

Under the regularity assumptions of \appref{sec:transfer_proof}, \cref{eq:prob_retention_angle} gives
\begin{equation}
 p_B(\theta(T))\ge
 \sin^2\!\left(\max\left\{0,
       \arcsin\sqrt{p_B(\theta(\tau))}-\tfrac12\mathcal P_{\tau:T}
                         \right\}\right).
 \label{eq:prob_retention_lower}
\end{equation}
For a fixed comparison probability $p_\star\in(0,1)$ under the same prompt law and success rule, a sufficient condition for $p_B(\theta(T))>p_\star$ is
\begin{equation}
 p_B(\theta(\tau))>p_\star,\qquad
 \mathcal P_{\tau:T}
 <2\left(\arcsin\sqrt{p_B(\theta(\tau))}-\arcsin\sqrt{p_\star}\right).
 \label{eq:prob_retained_advantage}
\end{equation}

\paragraph{Proof.}
Write $p_t=p_B(\theta(t))$ and $h_t=\langle\psi_0,\dot\theta(t)\rangle$, with $\psi_0$ evaluated at $\theta(t)$.
The score identity gives $\mathbb E[h_t]=0$ and $\dot p_t=\mathbb E[(o_B-p_t)h_t]$.
Hence
\begin{equation}
 |\dot p_t|\le
 \sqrt{p_t(1-p_t)}\sqrt{\mathbb E[h_t^2]}
 =\sqrt{p_t(1-p_t)}
  \sqrt{\dot\theta(t)^\top F(\theta(t))\dot\theta(t)}.
 \end{equation}
Since $\dot\alpha_t=\dot p_t/[2\sqrt{p_t(1-p_t)}]$, integration on each smooth segment and the triangle inequality prove \cref{eq:prob_retention_angle}.
Clipping its lower bound at zero and using monotonicity of $\sin^2$ on $[0,\pi/2]$ gives \cref{eq:prob_retention_lower} and \cref{eq:prob_retained_advantage}.
For sharpness, consider one prompt and two responses, with success probabilities $(\sin^2\alpha,\cos^2\alpha)$ and $\alpha\in(0,\pi/2)$.
The Fisher information for parameter $\alpha$ is $4$.
Every monotone path then has $\mathcal P_{\tau:T}=2|\alpha_T-\alpha_\tau|$, achieving equality.\hfill$\square$

For actual optimizer increments $\delta_j=\theta_{j+1}-\theta_j$, linear interpolation gives the exact path length
\begin{equation}
 \mathcal P_{\tau:T}
 =\sum_{j=\tau}^{T-1}\int_0^1
 \sqrt{\delta_j^\top F(\theta_j+s\delta_j)\delta_j}\,ds.
 \end{equation}
This form applies to a realized AdamW trajectory, including momentum and weight decay.

For the idealized mean-gradient path $\dot\theta=v_0+q\Delta v$ the Fisher seminorm satisfies $\sqrt{\dot\theta^\top F\dot\theta}\le\sqrt{v_0^\top Fv_0}+q\sqrt{\Delta v^\top F\Delta v}$, with every term evaluated at the evolving policy.

\section{Semantic Rubric Steering and the RubricOpt Loop}

\subsection{The Criterion Bank and Training-Only Conditioning}
\label{sec:prompts}
\Appref{sec:ontology} gives the frozen canonical names and Prefer/Avoid clauses used in formal training.
For each selected factor, the renderer takes its current clauses in the order $f_1$--$f_8$.
The main method selects all eight factors, while the ablation selects the groups in \cref{tab:ablation_detail}.
The renderer stays fixed across rounds:
\begin{quote}\ttfamily\small\raggedright
\textcolor{criterionhead}{Capability priorities:} <factor A>, <factor B>\\
\textcolor{criterionfavor}{Prefer:}\\
- <Prefer clause of factor A>;\\
- <Prefer clause of factor B>.\\
\textcolor{criterionavoid}{Avoid:}\\
- <Avoid clause of factor A>;\\
- <Avoid clause of factor B>.
\end{quote}
The Common rubric uses the $f_2$ and $f_5$ entries of \appref{sec:ontology}; larger sets extend the same lists.
With the Qwen3-4B tokenizer, Common, Common$+$Specific, Common$+$Low, and the full rubric cost $70$, $151$, $153$, and $234$ suffix tokens respectively.
Gate-accepted edits change clauses, not the factor order or renderer.

\paragraph{The Development Pilot.}
\label{sec:pilot}
The pilot compares ordinary DPO, a single-factor $f_2$ criterion, and a single-factor $f_7$ criterion while holding the pairs, labels, DPO objective, data order, and LoRA configuration fixed.
It trains Qwen3-4B for $150$ steps at a fixed conditioning rate $q=0.2$, so the $f_2$ and $f_7$ conditions condition exactly the same training rows and the comparison varies the criterion text alone.
Six Ascend 910B1 NPUs with per-device batch $2$ and accumulation $4$ give $48$ pairs per step and about $7{,}200$ pairs over those $150$ steps ($0.20$ epoch).
Maximum prompt and sequence lengths are $1{,}024$ and $2{,}048$ tokens, in bf16 with FlashAttention-2, and prompt truncation retains the final $1{,}024$ tokens, including the criterion suffix.
Each run takes about $47$ minutes.

Pilot evaluation uses its own sampling schedule. It draws $500$ prompts per benchmark with $4$ samples per prompt at temperature $0.7$, top-$p$ $0.8$, and top-$k$ $20$.
The unit of analysis is the prompt, and the paired tests are exact McNemar tests on the pass@4 indicator together with a two-sided sign-flip permutation test of $10{,}000$ resamples (seed $20260904$) on the paired per-prompt mean-correctness differences.
The two criteria differ only in their text and are within one token of each other in length.

Under the paired protocol, the $f_2$ criterion is favored over the $f_7$ criterion on all four pass@4 comparisons (\cref{tab:raw_pilot}).
The $f_2$ criterion is ahead on GSM8K and on MuSR pass@4 ($17$ passes unique to $f_2$ against $6$ unique to $f_7$), and MATH-Hard and BBH show the same direction.
The $f_7$ criterion shifts negatively relative to ordinary DPO on MuSR pass@4 ($3$ passes unique to $f_7$ against $14$ unique to ordinary DPO) and does not separate from ordinary DPO on the other three benchmarks.
The $f_2$ criterion and ordinary DPO do not differ on any of the four benchmarks, and ordinary DPO does not differ from the Instruct model on any of them.
Changing the criterion text alone therefore moved the two criteria apart on MuSR pass@4, and the $f_7$ criterion is the direction the diagnosis does not support.
Across the four benchmarks, counts of prompts passing only under the first versus only under the second condition are $[4/1,4/2,17/6,23/17]$ for $f_2$ versus $f_7$ and $[4/5,2/3,3/14,15/13]$ for $f_7$ versus ordinary DPO.

\begin{table}[H]
\centering
\tablestyle
\caption{Pilot accuracy and pass@4. Instruct is the untrained starting model. Trained conditions share pairs, labels, objective, data order, and LoRA configuration, differing only in criterion text. Bold marks row maxima, including ties.}
\label{tab:raw_pilot}
\footnotesize
\setlength{\tabcolsep}{5pt}
\begin{tabular}{lrrrr}
\toprule
\textbf{Benchmark} & \textbf{Instruct} & \textbf{Ordinary DPO} & \textbf{$f_2$ Criterion} & \textbf{$f_7$ Criterion}\\
\midrule
GSM8K accuracy & $0.8945$ & $0.8955$ & $\mathbf{0.9015}$ & $0.8910$\\
GSM8K pass@4 & $0.928$ & $0.928$ & $\mathbf{0.932}$ & $0.926$\\
\addlinespace[3pt]
MATH-Hard accuracy & $0.9070$ & $0.9105$ & $\mathbf{0.9125}$ & $0.9095$\\
MATH-Hard pass@4 & $0.964$ & $0.964$ & $\mathbf{0.966}$ & $0.962$\\
\addlinespace[3pt]
MuSR accuracy & $\mathbf{0.5870}$ & $0.5825$ & $0.5795$ & $0.5840$\\
MuSR pass@4 & $0.706$ & $\mathbf{0.708}$ & $\mathbf{0.708}$ & $0.686$\\
\addlinespace[3pt]
BBH accuracy & $0.0520$ & $0.0535$ & $\mathbf{0.0560}$ & $0.0495$\\
BBH pass@4 & $\mathbf{0.142}$ & $0.126$ & $\mathbf{0.142}$ & $0.130$\\
\bottomrule
\end{tabular}
\end{table}

\subsection{Stages, Contracts and the Deterministic Gate}
\label{sec:workflow_full}
\paragraph{Evidence Set and Engine.}
\Cref{fig:rubricopt} in the main text gives the six-step workflow.
The diagnostic stage reads item-level judge rationales for the all-eight-factor method against ordinary DPO and the Instruct model on Arena-Hard, MT-Bench (including a dedicated second-turn batch), and WildBench, together with generated correctness items carrying a gold answer on GSM8K, MATH-Hard, and MMLU-Pro.
That evidence is the first-round rubric-free benchmark evaluation output, read item by item, as described in \cref{sec:feedback_evaluation}.
The analysis and proposal engine is the current policy itself, the all-eight-factor merged checkpoint, so the model that reads the benchmark feedback is the model the feedback is about.
The external anchors are the gold answers and benchmark outcomes, the fixed judge's rationales and checklists, and the frozen $R_0$ text.

Five prompt stages process feedback from policy $\pi_{r+1}$, trained with the frozen round-$r$ rubric $R_r$, in the order 1 (benchmark diagnostic), 2 (cross-benchmark synthesis), 3 (factor attribution), 4 (conservative update), and 5 (rubric audit), followed by step 6 (deterministic gate).
Each stage is a separate call with its own schema, and no stage sees the next stage's output.

Stage 1 sees the factor schema, the criterion currently in force, and one benchmark's evidence, and returns a four-key YAML document:

\begin{quote}\ttfamily\small\raggedright
\textcolor{criterionhead}{preserve:}\\
\hspace*{1em}- behavior: / factor: / evidence:\\
\textcolor{criterionhead}{improve:}\\
\hspace*{1em}- failure\_mode: / factor: / evidence:\\
\textcolor{criterionhead}{criterion\_gaps:}\\
\hspace*{1em}- factor: / current\_problem: / suggested\_direction:\\
\textcolor{criterionhead}{overall\_diagnosis:}
\end{quote}

Stage 1 uses only $f_1$--$f_8$ and separates behaviors to preserve from those to improve.
An isolated example requires strong evidence, and attributing a benchmark change to the criterion requires behavioral evidence.
The output must be concrete and concise.

Stage 2 aggregates every stage-1 report for one policy.
It may not rewrite the criterion or propose wording.
It organises evidence and records disagreement, adding the benchmarks that support each pattern, an evidence-strength level (strong, moderate, weak), and an explicit record of contradictions.
A benchmark where the policy did not change still counts as evidence about what is stable.

Stage 3 attributes each recurring behaviour onto the ontology and decides whether the current wording already covers it.
The coverage decision follows this rule:

\begin{quote}
If the current Prefer/Avoid wording states this failure mode in full, the model simply did not learn it: the decision is \textsc{keep}.
If the wording does not state the failure mode, or states only part of it, \emph{that is the only case in which \textup{\textsc{refine}} may be considered}.
\end{quote}

Attribution disambiguates reasoning validity, verification and calibration, completeness, and instruction adherence, naming the primary factor when several are involved.
It attributes only what the evidence supports.

Stage 4 proposes the smallest clause edits that address high-confidence recurring gaps while preserving factor identity and behaviors marked for preservation.
Edits may clarify wording, narrow an overly broad reading, make an implicit requirement explicit, or refine a matched Avoid clause.
Stage 4 prohibits examples, task-specific tricks, benchmark or dataset names, scores, numerical weights, and intensifiers.
Unsupported changes return \textsc{keep}.
The output is a candidate $R_{r+1}$ and a per-factor ledger with the action, reason, and old and new clauses.

Stage 5 audits semantic identity, benchmark and dataset leakage, evidential support, specificity, Prefer/Avoid matching, preserved behaviors, and edit magnitude.
It emits a justification per axis and one \texttt{PASS} or \texttt{FAIL}.
The audit is advisory rather than decisive: in one executed round it returned a clean eight-of-eight \texttt{PASS} on a candidate that the deterministic gate then rejected.

\paragraph{Step 6: The Deterministic Gate.}
The binding checks constrain eligibility, semantic identity, and procedural consistency.

Acceptance is decided by eight checks in round 1 and ten in round 2.
Round 2 adds that only gate-accepted factors may differ, and that the recorded decisions are consistent with the rendered rubric.

\begin{quote}\ttfamily\small\raggedright
\begin{enumerate}\itemsep2pt
\item All eight factors must occur exactly once, each with a \textsc{keep}/\textsc{refine} decision.
\item The input clauses of every refinement must match the frozen canonical wording byte-for-byte.
\item Each new clause must retain at least $0.60$ of the old clause's content tokens, with its length change inside $[-5,+10]$ words.
\item Every factor must retain one matched Prefer/Avoid pair.
\item Clauses must contain no benchmark or dataset names, digits, intensifiers, or example markers.
\item Every refinement must be traceable to an attribution judging the failure not fully covered.
\item Edit magnitude must remain within the specified budget.
\item Cross-factor vocabulary imports raise a warning rather than a rejection.
\item In round 2, only gate-accepted factors may differ.
\item In round 2, the recorded decisions must agree with the rendered rubric.
\end{enumerate}
\end{quote}

Failures of the binding checks reject the affected edit, so a factor that violates a check is reverted to \textsc{keep}.

\subsection{Matched Re-Optimization and the Dataset Control}
\label{sec:r2_protocol}
The shared recipe is in \cref{sec:whatchanges}: rank-$64$ LoRA ($\alpha=128$, dropout $0.05$), mean-per-token DPO with $\beta=1$ at learning rate $10^{-6}$, global batch $64$, and seed $42$.
The formal runs and the HelpSteer3 continuations take $566$ steps.
The HelpSteer3 continuations start from the frozen \Pone{} checkpoint and use fresh AdamW and FP32 LoRA adapters, cosine decay with $5\%$ warmup, and one shared exposure mask on identical ordered pairs.
Their reference is the frozen \Pone{} checkpoint with the new adapter disabled, and the starting policy of every other block is the released Instruct checkpoint with its own adapter disabled.
Only the rubric text differs: \textsc{keep} uses $R_0$ and \textsc{refine} uses $R_1$.
The primary comparison \PtwoR{} minus \PtwoK{} measures editing; comparison with \Pone{} includes further training.

Tulu3 uses the same initial rubric and training recipe on an independently assembled dataset, over $1{,}818$ steps at global batch $64$.
The budgets match ordinary DPO within each dataset by epoch count, not across datasets by step count.
The Tulu3 second-round change combines editing and additional training.

\section{Experimental Settings and Results}

This appendix gives the measurement and protocol detail behind the main-text results: evaluation protocols, training-protocol audits, and the per-cell measurements that the main tables summarize.

\subsection{Evaluation and Measurement Protocols}
\label{sec:setup_details}

\paragraph{Splits and Evaluation Protocols.}
The factor measurements use diagnostic and held-out subsets assigned by an \texttt{md5} hash of the prompt identifier, and comparing these subsets tests the stability of the measured demand gaps.
Evaluation prompts come from benchmark sources, while the preference-training pairs come from HelpSteer3.
Rubric refinement uses first-round benchmark evaluation outputs as feedback and compares the continuation branches on matched evaluation items (\cref{sec:feedback_evaluation}).
The formal study uses the released benchmarks through the standard evaluation harness, with adapter evaluation enabled and the adapter merged, and the reported baselines use this protocol (\cref{tab:main_results,tab:ablation_detail}).
The demand values in \cref{tab:demand_detail} are for Qwen3-4B.

\paragraph{Open-Ended Protocols.}
Arena-Hard and WildBench use greedy decoding with a $4{,}096$-token generation limit and one response per prompt.
WildBench uses $256$ prompts and MiniMax-M3 scores on a $1$--$10$ scale, averaged over usable verdicts ($255$ for Qwen3-8B \Pone{}, $256$ for the other reported conditions).
MT-Bench uses a $1{,}024$-token generation limit.
Every Arena-Hard condition is judged against the fixed answers of the standard Arena-Hard anchor (GPT-4-0314), with the candidate always in slot B, so Arena-Hard here is a \emph{fixed-anchor win rate}: every method faces the same anchor on the same items, rather than a head-to-head between methods.
Per-item outcomes are wins, ties, or losses, with ties counted as half-wins, and all conditions use the same tie rule, anchor, and items.

\paragraph{Accuracy Evaluation Protocols.}
\label{sec:ev_reeval}
Absolute scores are reported in \cref{tab:main_results,tab:mistral_results,tab:continuation_main,tab:ablation_detail}.
BBH covers all $27$ subtasks in the project's sampled evaluation.
The paired differences in \cref{tab:ev_paired_final} use the same $(\mathrm{prompt\_id},\mathrm{sample\_idx})$ keys within each model--dataset cell.
MATH-Hard uses the lm-eval task template with $7$ subtasks.
Scores are the unweighted mean exact match over subtasks under vLLM/bf16 with merged adapters.

\paragraph{Training-Protocol Details.}
The trainer computes \cref{eq:objective} with each response's own unpadded token count.
Policy and frozen reference receive identical batch tensors and context; reference log-probabilities are recomputed with the adapter disabled and no cache.
Rubric tokens remain in the masked prompt region and enter neither response-token numerator nor denominator, and the chosen and rejected response text is unchanged by conditioning.

\paragraph{Matched-Length Criterion Comparison.}
At formal scale, Common$+$Specific and Common$+$Low vary the criterion text, and different additions to Common.
Common$+$Specific exceeds Common$+$Low by $0.07$ BBH points, $0.01$ points on the seven-task average, and $0.5$ Arena-Hard points.

\paragraph{Formal Runs.}
Formal runs use eight Ascend NPUs under distributed data parallel.
The realized conditioning exposure is $0.5009$ on HelpSteer3 and $0.4971$ on the Tulu3 conditioned runs, and the criterion is withdrawn for the final $10\%$ of training and for main evaluation.

\paragraph{Inference-Time Rubric Reinsertion.}
\label{sec:reinsertion_protocol}
The same Qwen3-4B \Ptwo{} checkpoint is evaluated with no rubric, the trained wording $R_1$, and the untrained candidate $R_2$.
The rubric is inserted at the last message before templating, adding $243$ and $256$ prompt tokens respectively.
\Cref{tab:ev_testA} reports six benchmarks generated in one matched batch window, with identical $(\mathrm{prompt\_id},\mathrm{sample\_idx})$ sets of $2{,}000$ rows per benchmark.
On MATH-Hard, OFF scores $63.17\%$, $R_1$ ON $63.85\%$, and $R_2$ ON $63.93\%$.

\begin{table}[!htbp]
\centering
\tablestyle
\caption{Rubric reinsertion on the same checkpoint and matched sampled items (accuracy fractions). }
\label{tab:ev_testA}
\footnotesize
\setlength{\tabcolsep}{4pt}
\begin{tabular}{lrrrrrr}
\toprule
\textbf{Condition} & \textbf{GSM8K} & \textbf{MMLU-Pro} & \textbf{TruthfulQA} & \textbf{ARC} & \textbf{BBH} & \textbf{MuSR}\\
\midrule
OFF & $0.8995$ & $\mathbf{0.6390}$ & $\mathbf{0.6955}$ & $\mathbf{0.9070}$ & $0.4310$ & $0.5755$\\
$R_1$ ON & $0.8960$ & $0.6205$ & $0.6895$ & $0.8945$ & $0.4185$ & $0.5640$\\
$R_2$ ON & $\mathbf{0.9055}$ & $0.6285$ & $0.6905$ & $0.9000$ & $\mathbf{0.4680}$ & $\mathbf{0.5775}$\\
\bottomrule
\end{tabular}
\end{table}

\paragraph{The Demand Matrices.}
\label{sec:demand_detail}
\Cref{tab:demand_detail} reports the full Qwen3-4B demand matrix with effect sizes.
The largest diagnostic--held-out gap difference is $0.045$, with no sign reversal.
The 8B matrix is in \cref{tab:demand_8b}.

Factual and evidential grounding separates success from failure more strongly on the knowledge benchmarks than on the reasoning benchmarks ($0.111$ against $0.031$, \cref{fig:diagnosis}b).
Within the knowledge group, ARC is driven by reasoning validity, TruthfulQA by verification and factual grounding, and MMLU-Pro by completeness.
MuSR is close to flat on every factor.
A deployment target usually contains several benchmarks with different semantic needs, and how they are combined changes which factor appears most important. On the knowledge group the equal-per-benchmark average puts reasoning validity first ($0.213$) and completeness second ($0.193$), whereas pooling by response count reverses the order, because MMLU-Pro supplies about $86\%$ of the responses in the group, raising completeness to $0.356$.
The per-benchmark profiles are kept. The equal-per-benchmark average is a descriptive summary.

\begin{table}[!htbp]
\centering
\tablestyle
\caption{Qwen3-4B diagnostic matrix: mean success--failure gaps on the $0$--$4$ judge scale, followed by Cohen's $d$. Bold shaded gaps are at least $0.15$.}
\label{tab:demand_detail}
\footnotesize
\setlength{\tabcolsep}{4pt}
\begin{tabular}{lrrrrrrr}
\toprule
\multicolumn{8}{l}{\bfseries Mean factor-score gap}\\
\midrule
\textbf{Factor} & \textbf{GSM8K} & \textbf{MATH-Hard} & \textbf{BBH} & \textbf{MuSR} & \textbf{ARC} & \textbf{TruthfulQA} & \textbf{MMLU-Pro}\\
\midrule
$f_1$ & $+0.080$ & $+0.039$ & $+0.023$ & $+0.005$ & $+0.110$ & $+0.100$ & $+0.069$\\
$f_2$ & \cellcolor{tablerubric}$\mathbf{+0.170}$ & \cellcolor{tablerubric}$\mathbf{+0.246}$ & $+0.066$ & $+0.047$ & \cellcolor{tablerubric}$\mathbf{+0.248}$ & \cellcolor{tablerubric}$\mathbf{+0.157}$ & \cellcolor{tablerubric}$\mathbf{+0.235}$\\
$f_3$ & $-0.051$ & $+0.090$ & $+0.018$ & $+0.025$ & $+0.105$ & $+0.145$ & $+0.142$\\
$f_4$ & $+0.036$ & $+0.054$ & $+0.028$ & $+0.006$ & $+0.093$ & \cellcolor{tablerubric}$\mathbf{+0.150}$ & $+0.091$\\
$f_5$ & \cellcolor{tablerubric}$\mathbf{+0.158}$ & \cellcolor{tablerubric}$\mathbf{+0.486}$ & \cellcolor{tablerubric}$\mathbf{+0.232}$ & $+0.038$ & $+0.148$ & $+0.034$ & \cellcolor{tablerubric}$\mathbf{+0.399}$\\
$f_6$ & $+0.086$ & $+0.065$ & $+0.075$ & $+0.010$ & $-0.007$ & $+0.035$ & $+0.038$\\
$f_7$ & $-0.048$ & \cellcolor{tablerubric}$\mathbf{+0.164}$ & \cellcolor{tablerubric}$\mathbf{+0.235}$ & $+0.069$ & $+0.040$ & $-0.020$ & $+0.132$\\
$f_8$ & $+0.081$ & \cellcolor{tablerubric}$\mathbf{+0.162}$ & $+0.095$ & $+0.069$ & \cellcolor{tablerubric}$\mathbf{+0.179}$ & $+0.110$ & \cellcolor{tablerubric}$\mathbf{+0.219}$\\
\midrule
\multicolumn{8}{l}{\bfseries Cohen's $d$}\\
\midrule
\textbf{Factor} & \textbf{GSM8K} & \textbf{MATH-Hard} & \textbf{BBH} & \textbf{MuSR} & \textbf{ARC} & \textbf{TruthfulQA} & \textbf{MMLU-Pro}\\
\midrule
$f_1$ & $+0.578$ & $+0.326$ & $+0.103$ & $+0.021$ & $+0.680$ & $+0.280$ & $+0.275$\\
$f_2$ & $+0.756$ & $+0.709$ & $+0.160$ & $+0.123$ & $+0.988$ & $+0.362$ & $+0.485$\\
$f_3$ & $-0.112$ & $+0.167$ & $+0.035$ & $+0.050$ & $+0.210$ & $+0.288$ & $+0.283$\\
$f_4$ & $+0.319$ & $+0.313$ & $+0.124$ & $+0.049$ & $+0.570$ & $+0.327$ & $+0.256$\\
$f_5$ & $+0.588$ & $+1.105$ & $+0.408$ & $+0.069$ & $+0.297$ & $+0.065$ & $+0.700$\\
$f_6$ & $+0.270$ & $+0.267$ & $+0.200$ & $+0.038$ & $-0.023$ & $+0.081$ & $+0.122$\\
$f_7$ & $-0.093$ & $+0.315$ & $+0.447$ & $+0.147$ & $+0.090$ & $-0.043$ & $+0.278$\\
$f_8$ & $+0.222$ & $+0.366$ & $+0.196$ & $+0.136$ & $+0.317$ & $+0.232$ & $+0.387$\\
\bottomrule
\end{tabular}
\end{table}

\paragraph{Model-Dependent Capability Signatures.}
Demand changes with the evaluated model as well as with the task.
The Qwen3-4B and Qwen3-8B profiles correlate strongly on TruthfulQA ($r=0.953$), GSM8K ($r=0.906$), and MMLU-Pro ($r=0.868$), moderately on MuSR ($r=0.522$), MATH-Hard ($r=0.461$), and ARC ($r=0.457$), and negatively on BBH ($r=-0.751$). The disagreement is not confined to one factor (\cref{fig:model_demand_slopes}a): on BBH, $f_5$, $f_7$, $f_6$, and $f_2$ all change sign. The $f_5$ gap changes from $+0.232$ on Qwen3-4B to $-0.100$ on Qwen3-8B, and the $f_7$ gap from $+0.235$ to $-0.171$.
The demand scope changes as well, and that matters more for a diagnosis than any single gap: reasoning validity is meaningful on five of the seven benchmarks for Qwen3-4B but on only one for Qwen3-8B, verification and calibration rises from none to two, and clarity and organization remains meaningful on two (\cref{tab:demand_8b}).
A priority read off the weaker model would not describe the stronger one, which is consistent with a fixed ontology and a changing capability signature rather than one universal factor ranking, and it makes the grouping of \cref{sec:diagnosis} specific to Qwen3-4B.

\paragraph{Demand and Supply.}
Reasoning validity ($f_2$) is meaningful on five benchmarks but has the lowest coverage ($0.146$), while completeness ($f_5$) is meaningful on four and has the highest coverage ($0.474$).
The length controls below show that their supply contrasts have different sensitivity to response length.

Chosen responses are longer, so we also report a length-matched subset: pairs whose chosen and rejected responses differ by at most $164$ characters, the lower-quartile bound of the length-difference distribution, giving $n=9{,}551$.
For each factor we additionally fit $\Delta z_k=\alpha+\gamma\,z(\Delta L)$ over the full set and report $\gamma$ per standard deviation of the length difference.
Across the factors we examined, $\gamma$ is $0.040$ for reasoning validity, $0.021$ for factual grounding, $0.175$ for completeness, $0.105$ for clarity, and $-0.177$ for efficiency.

\paragraph{Supply Detail and Length Matching.}
\Cref{tab:ev_supply} reports mean contrasts, coverage, positive fractions, and tie rates.
Negative contrasts are a minority for every factor (at most $0.171$), the contrast SDs lie between $0.476$ and $0.835$, and Cohen's $d_z$ between $0.089$ and $0.193$.

\begin{table}[!htbp]
\centering
\tablestyle
\caption{HelpSteer3 supply statistics over all assessable pairs: mean signed contrast $\mu_k$, contrast coverage $C_k=\Pr(|\delta|\geq1)$, and positive fraction $P_k=\Pr(\delta>0)$. Shading marks $C_k<0.25$. Reduction is the relative percentage drop in mean contrast after length matching, where reported.}
\label{tab:ev_supply}
\footnotesize
\begin{tabular}{lrrrrl}
\toprule
\textbf{Factor} & \textbf{$\mu_k$} & \textbf{$C_k$} & \textbf{$P_k$} & \textbf{Tie rate} & \textbf{Reduction}\\
\midrule
$f_1$ task comprehension & $0.1042$ & $0.3379$ & $0.2118$ & $0.6621$  & ---\\
$f_2$ reasoning validity & $0.0908$ & \cellcolor{tableflag}$\mathbf{0.1463}$ & $0.1032$ & $\mathbf{0.8537}$  & $13\%$\\
$f_3$ verification and calibration & $0.0818$ & \cellcolor{tableflag}$0.2202$ & $0.1506$ & $0.7798$  & ---\\
$f_4$ factual and evidential grounding & $0.0866$ & \cellcolor{tableflag}$\mathbf{0.1493}$ & $0.0992$ & $0.8507$  & ---\\
$f_5$ completeness & $0.1556$ & $\mathbf{0.4743}$ & $0.3037$ & $0.5257$  & $38\%$\\
$f_6$ instruction adherence & $0.0988$ & $0.3405$ & $0.2046$ & $0.6595$  & ---\\
$f_7$ clarity and organization & $0.1428$ & $0.3950$ & $0.2611$ & $0.6050$  & $47\%$\\
$f_8$ efficiency and conciseness & $0.0657$ & $0.3248$ & $0.1850$ & $0.6752$  & ---\\
\bottomrule
\end{tabular}
\end{table}

\begin{figure}[!htbp]
\centering
\includegraphics[width=\linewidth]{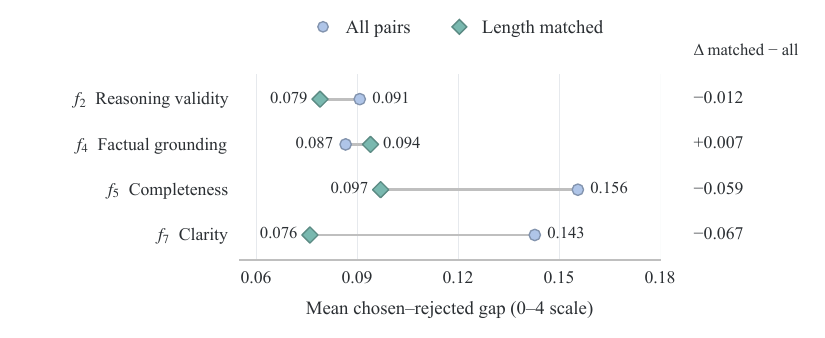}
\caption{Length sensitivity of HelpSteer3 supply estimates. Points show mean chosen--rejected gaps for four factors on the $0$--$4$ judge scale, using all $38{,}173$ pairs or $9{,}551$ length-matched pairs ($|\Delta L|\leq164$ characters). $\Delta$ is matched minus all.}
\label{fig:length_control}
\end{figure}

The $38{,}173$ assessable pairs include $1{,}917$ held out from training, and the same coverage values are reused in the 8B diagnosis.
All factors have more positive than negative contrast mass, but $f_2$ and $f_4$ tie on about $85\%$ of pairs.
$59.2\%$ of pairs have a longer chosen response, with mean $+180$ and median $+106$ characters.
On the length-matched subset ($|\Delta L|\leq164$, $n=9{,}551$, \cref{fig:length_control}), mean contrasts change from $0.156$ to $0.097$ for $f_5$, $0.143$ to $0.076$ for $f_7$, $0.091$ to $0.079$ for $f_2$, and $0.087$ to $0.094$ for $f_4$.
Among these four factors, reasoning validity and factual grounding change less than completeness and clarity.
The length slope is $\gamma=+0.175$ for $f_5$, $+0.105$ for $f_7$, $-0.177$ for $f_8$, $0.040$ for $f_2$, and $0.021$ for $f_4$.
This control assesses the length sensitivity of the supply diagnostic; the effect of rubric self-revision is evaluated by the matched continuations in \cref{tab:continuation_main}.

\subsection{Paired Effects and Results Across Datasets and Models}
\label{sec:paired}
\Cref{tab:ev_paired_final} reports BBH contrasts from the full-subtask evaluation.
MATH-Hard changes use the separate greedy protocol of \appref{sec:ev_reeval}.

\begin{table}[!htbp]
\centering
\tablestyle
\caption{Full-subtask BBH differences within each model--dataset cell (accuracy fractions).  Shading marks the matched \textsc{refine}-minus-\textsc{keep} comparison.}
\label{tab:ev_paired_final}
\footnotesize
\setlength{\tabcolsep}{4pt}
\begin{tabular}{llrl}
\toprule
\textbf{Cell} & \textbf{Comparison} & \textbf{$\Delta$} & \textbf{Reading}\\
\midrule
4B $\times$ HS3 & Common $-$ Ordinary DPO & $+0.0090$ & positive\\
 & Common$+$Specific $-$ Ordinary DPO & $+0.0047$ & positive\\
 & Common$+$Low $-$ Ordinary DPO & $+0.0040$ & positive\\
 & \Pone{} $-$ Ordinary DPO & $+0.0023$ & null\\
 & \PtwoK{} $-$ \Pone{} & $+0.0309$ & positive\\
\rowcolor{tablerubric}
 & \PtwoR{} $-$ \PtwoK{} & $-0.0025$ & null\\
\midrule
4B $\times$ Tulu3 & \Pone{} $-$ Ordinary DPO & $+0.0420$ & positive\\
\midrule
8B $\times$ Tulu3 & \Pone{} $-$ Ordinary DPO & $+0.0961$ & positive\\
\bottomrule
\end{tabular}
\end{table}

\begin{table}[!ht]
\centering
\tablestyle
\caption{Per-benchmark factor-set results on Qwen3-4B with HelpSteer3. Accuracy and Arena-Hard win rate are percentages. Avg.\ is the unweighted seven-task accuracy mean. Open-ended protocols use native scores. $\mathrm{Avg.}\,\Delta$ is the relative change against ordinary DPO across MT-Bench, Arena-Hard, and WildBench. Each rubric includes Common, with Specific and/or Low-demand factors added. Bold marks column maxima. Shading marks rubric-trained rows.}
\label{tab:ablation_detail}
\setlength{\tabcolsep}{1.2pt}
\fontsize{7.5}{9}\selectfont
\begin{tabular}{lrrrrrrrrrrrr}
\toprule
& \multicolumn{8}{c}{\shortstack{\textbf{HelpSteer3}\\\textbf{comprehensive capability (\%)}}} & \multicolumn{4}{c}{\shortstack{\textbf{HelpSteer3}\\\textbf{open-ended generation}}}\\
\cmidrule(lr){2-9}\cmidrule(lr){10-13}
\textbf{Method} & \textbf{GSM8K} & \textbf{\shortstack{MATH-\\Hard}} & \textbf{BBH} & \textbf{MuSR} & \textbf{ARC} & \textbf{\shortstack{Truthful\\QA}} & \textbf{\shortstack{MMLU-\\Pro}} & \textbf{Avg.} & \textbf{\shortstack{MT-\\Bench}} & \textbf{\shortstack{Arena-\\Hard}} & \textbf{\shortstack{Wild-\\Bench}} & \textbf{\shortstack{Avg.\\$\Delta$}}\\
\midrule
\tablepanel{13}{Qwen3-4B}
Instruct & $85.90$ & $62.24$ & $12.82$ & $\mathbf{43.39}$ & $62.80$ & $52.27$ & $41.98$ & $51.63$ & $6.320$ & $49.0\%$ & $3.730$ & $-2.96$\\
Ordinary DPO & $\mathbf{87.57}$ & $63.64$ & $14.73$ & $43.12$ & $62.63$ & $\mathbf{53.91}$ & $\mathbf{42.30}$ & $52.56$ & $6.890$ & $50.9\%$ & $3.617$ & $0.00$\\
\rowcolor{tablerubric}
Common & $87.04$ & $63.48$ & $\mathbf{15.64}$ & $43.12$ & $63.23$ & $53.72$ & $42.23$ & $52.64$ & $6.720$ & $52.3\%$ & $3.711$ & $+0.96$\\
\rowcolor{tablerubric}
Common$+$Specific & $87.04$ & $63.74$ & $15.21$ & $42.99$ & $\mathbf{63.40}$ & $53.71$ & $42.25$ & $52.62$ & $6.610$ & $\mathbf{53.1\%}$ & $\mathbf{3.840}$ & $+2.14$\\
\rowcolor{tablerubric}
Common$+$Specific$+$Low & $87.49$ & $\mathbf{64.18}$ & $14.96$ & $43.25$ & $63.31$ & $53.74$ & $42.25$ & $\mathbf{52.74}$ & $\mathbf{6.930}$ & $53.0\%$ & $3.727$ & $\mathbf{+2.58}$\\
\rowcolor{tablerubric}
Common$+$Low & $86.88$ & $64.02$ & $15.14$ & $43.12$ & $63.14$ & $53.69$ & $42.26$ & $52.61$ & $6.900$ & $52.6\%$ & $3.602$ & $+1.02$\\
\bottomrule
\end{tabular}
\color{black}
\end{table}

\paragraph{Per-Benchmark Differences and the Matched Continuation.}
Five comparisons are fixed in advance, each holding pairs, labels, objective, budget, and schedule fixed: A (Common minus ordinary DPO) asks whether the broadly demanded factors help, B (Common $+$ Specific minus Common) measures the additional specific factors, C (Common $+$ Low minus Common) measures the additional low-demand factors, D (full rubric minus Common $+$ Specific) measures adding Low once Specific is present, and E (Common $+$ Specific minus Common $+$ Low) compares the two additions.
The absolute scores for these comparisons are in \cref{tab:ablation_detail}.
Reported per-benchmark differences range from $-0.53$ to $+0.90$ percentage points.
The corresponding changes in the unweighted seven-benchmark mean are $+0.08$, $-0.02$, $-0.03$, $+0.12$, and $+0.01$ points for A--E, respectively.
\Cref{tab:continuation_main} reports the starting policy and both HelpSteer3 continuations.
The shared configuration in \appref{sec:r2_protocol} makes \PtwoR{} minus \PtwoK{} the edit comparison.
Besides the main-text BBH and MT-Bench changes, the edit changes MATH-Hard by $-0.10$ points and WildBench by $+0.055$ native-score units.

The Qwen3 and Mistral benchmark suites are in \cref{tab:main_results,tab:mistral_results}; factor-set results are in \cref{tab:ablation_detail}.
\Pbase{} is the released checkpoint every method starts from, \Pdpo{} is rubric-free preference optimisation on the same pairs, and \Pone{} is the policy after the first RubricOpt round, trained under the initial rubric $R_0$ over all eight factors.
\PtwoK{} and \PtwoR{} are the two matched second-round branches: both start from the same frozen \Pone, run one epoch over the same pairs with the same exposure mask and optimiser, and differ only in whether the rubric is the unchanged $R_0$ or the refined $R_1$.

\paragraph{The Second Dataset.}
The Tulu3 refined continuation starts from the merged \Pone{} policy and changes $f_3$, $f_4$, and $f_8$.
Its absolute scores are in \cref{tab:main_results}, and the continuation changes MATH-Hard accuracy from $43.35\%$ to $37.12\%$, a $-6.23$-point joint change from additional training and editing.

\paragraph{The Second Model.}
\Cref{tab:demand_detail,tab:demand_8b} report the demand gaps for the two scales.
The 8B block trains an ordinary-DPO method and a \Pone{} method on the first dataset, $566$ steps each from the released checkpoint at global batch $64$ with the same LoRA and seed, differing only in the rubric block.
Both methods are evaluated on all seven benchmarks, and the conditioned method has $14{,}000$ factor-scored responses.

\begin{table}[!htbp]
\centering
\tablestyle
\caption{Qwen3-8B success--failure gaps on the $0$--$4$ judge scale, pooling diagnostic and held-out halves within the model. Bold shaded cells have $|\mathrm{gap}|\geq0.15$. The Qwen3-4B matrix is in \cref{tab:demand_detail}.}
\label{tab:demand_8b}
\footnotesize
\setlength{\tabcolsep}{3pt}
\begin{tabular}{lrrrrrrrr}
\toprule
\textbf{Benchmark} & \textbf{$f_1$} & \textbf{$f_2$} & \textbf{$f_3$} & \textbf{$f_4$} & \textbf{$f_5$} & \textbf{$f_6$} & \textbf{$f_7$} & \textbf{$f_8$}\\
\midrule
GSM8K & $+0.113$ & $+0.130$ & $-0.096$ & $+0.014$ & \cellcolor{tablerubric}$\mathbf{+0.189}$ & $+0.058$ & $-0.065$ & \cellcolor{tablerubric}$\mathbf{+0.171}$ \\
MATH-Hard & $+0.011$ & $+0.079$ & \cellcolor{tablerubric}$\mathbf{+0.182}$ & $+0.013$ & $+0.144$ & $+0.058$ & \cellcolor{tablerubric}$\mathbf{+0.163}$ & $+0.080$\\
BBH & $-0.038$ & $-0.007$ & $+0.142$ & $+0.001$ & $-0.100$ & $-0.028$ & \cellcolor{tablerubric}$\mathbf{-0.171}$ & $+0.073$\\
MuSR & $+0.022$ & $+0.042$ & $+0.039$ & $+0.012$ & $+0.042$ & $+0.022$ & $+0.038$ & $+0.024$\\
ARC & $+0.072$ & \cellcolor{tablerubric}$\mathbf{+0.167}$ & $+0.109$ & $+0.089$ & $+0.141$ & $+0.033$ & $-0.006$ & $-0.035$\\
TruthfulQA & $+0.016$ & $+0.053$ & $+0.078$ & $+0.100$ & $-0.017$ & $-0.031$ & $-0.131$ & $+0.039$\\
MMLU-Pro & $+0.057$ & $+0.136$ & \cellcolor{tablerubric}$\mathbf{+0.190}$ & $+0.059$ & \cellcolor{tablerubric}$\mathbf{+0.521}$ & $+0.092$ & \cellcolor{tablerubric}$\mathbf{+0.183}$ & $+0.137$\\
\bottomrule
\end{tabular}
\end{table}

\vspace{-2mm}
\paragraph{Mistral Evaluation.}
\label{sec:ev_mistral}
Mistral-7B uses the same recipe, with Tulu3 MATH-Hard and BBH read through the extraction ladder rather than the Qwen3 protocols.
\Cref{tab:mistral_results} reports both suites.
BBH falls by $7.95$ points against its own ordinary-DPO control, with the seven-task average unchanged.

\end{document}